\documentclass[letterpaper]{article} 
\usepackage{aaai25}  
\usepackage{times}  
\usepackage{helvet}  
\usepackage{courier}  
\usepackage[hyphens]{url}  
\usepackage{graphicx} 
\usepackage{natbib}  
\usepackage{caption} 
\usepackage{forest}

\usepackage{fancyvrb} 
\usepackage{times}
\usepackage{soul}
\usepackage{url}
\usepackage{graphicx}
\usepackage{amsmath}
\usepackage{amsthm}
\usepackage{booktabs}
\usepackage{algorithm}
\usepackage{xcolor}
\usepackage{amsfonts}
\usepackage[noend]{algpseudocode}
\usepackage{subcaption}
\usepackage{footnotehyper}

\newcommand{\hide}[1]{}

\newtheorem{defn}{Definition}
\newtheorem{example}{Example}

\usepackage{newfloat}
\usepackage{listings}
\DeclareCaptionStyle{ruled}{labelfont=normalfont,labelsep=colon,strut=off} 
\floatstyle{ruled}
\newfloat{listing}{tb}{lst}{}
\floatname{listing}{Listing}
\title{Generalizing Constraint Models in Constraint Acquisition}

\author{
    Dimos Tsouros\textsuperscript{\rm 1}, 
    Senne Berden\textsuperscript{\rm 1}, 
    Steven Prestwich\textsuperscript{\rm 2}
    Tias Guns\textsuperscript{\rm 1}
}
\affiliations{
    \textsuperscript{\rm 1}Department of Computer Science, KU Leuven, Belgium\\
    \textsuperscript{\rm 2}School of Computer Science and Information Technology, University College Cork\\

    dimos.tsouros@kuleuven.be, senne.berden@kuleuven.be, s.prestwich@cs.ucc.ie, tias.guns@kuleuven.be
}

\usepackage{bibentry}

\begin{document}

\maketitle

\begin{abstract}
Constraint Acquisition (CA) aims to widen the use of constraint programming by assisting users in the modeling process. However, most CA methods suffer from a significant drawback: they learn a single set of individual constraints for a specific problem instance, but cannot generalize these constraints to the parameterized constraint specifications of the problem. In this paper, we address this limitation by proposing \textsc{GenCon}, a novel approach to learn parameterized constraint models capable of modeling varying instances of the same problem.
To achieve this generalization, we make use of statistical learning techniques at the level of individual constraints.
Specifically, we propose to train a classifier to predict, for any possible constraint and parameterization, whether the constraint belongs to the problem.
We then show how, for some classes of classifiers, we can extract decision rules to construct interpretable constraint specifications. This enables the generation of ground constraints for any parameter instantiation.
Additionally, we present a generate-and-test approach that can be used with any classifier, to generate the ground constraints on the fly. 
Our empirical results demonstrate that our approach achieves high accuracy and is robust to noise in the input instances.
\end{abstract}


 \begin{links}
     \link{Code}{https://github.com/Dimosts/GenConModels}
 \end{links}

\section{Introduction}

Constraint Programming (CP) is considered one of the main paradigms for solving combinatorial problems in AI. It provides powerful modeling languages and solvers for decision-making, with many successful applications~\cite{wallace1996practical,simonis1999building}. 
In CP, the user declaratively states the constraints over a set of decision variables, thereby defining the feasible solutions to their problem. A solver is then used to generate a solution. However, modeling a new application as a constraint problem requires significant expertise, which is 
a barrier to the wider use of CP~\cite{freuder2014grand,freuder2018progress}. This has motivated the development of methods to assist the user in the modeling process~\cite{de2018learning,freuder2018progress,kolb2016learning,lombardi2018boosting}. This is the focus of the research area of \textit{Constraint Acquisition (CA)}~\cite{bessiere2017constraint}, which has been identified as an important topic for CP~\cite{de2018learning} and as progress toward the ``Holy Grail'' of computer science~\cite{freuder2018progress}.

In CA, constraints are either learned passively from a set of known solutions and (optionally) non-solutions~\cite{beldiceanu2012model,berden2022learning,prestwich2021classifier} or actively through interaction with a user~\cite{bessiere2023learning,tsouros2020efficient,tsouros2021learning}. Recent advancements in both passive and active acquisition systems show significant potential~\cite{prestwich2021classifier,prestwich2020robust,tsouros2023learning}.
For instance, a recent application of interactive CA in a real-world scheduling problem was presented in~\cite{barral2024}.

However, a significant limitation of most (passive and active) CA systems is that they only learn the ground constraints of a specific problem instance~\cite{ArcEtc,bessiere2017constraint,Pre2,prestwich2021classifier,TsoEtc}, while in practice it is common for the instance at hand to change over time. To accommodate these changes, well-known constraint modeling languages like MiniZinc~\cite{minizinc} and CPMpy~\cite{guns2019increasing} allow the use of parameters in the constraint model.
An assignment of values to the parameters then instantiates the ground constraints for a given instance of the problem.
For example, in exam timetabling with the requirement `exams of courses that are given in the same semester should be scheduled on different days', the actual ground constraints will be instantiated based on the following parameters: which semester each course belongs to, and how many timeslots are available per day. Achieving the same ability to \textit{generalize} the constraint models learned using CA over different instances of the same problem class 
has been identified as a key challenge for CA~\cite{simonis2023requirements}.

In the literature on CA, there are only a few works that support generalization.
The first approach for generalizing constraints was introduced in interactive CA~\cite{bessiere2014boosting,daoudi2015detecting}, to enhance the acquisition process within a single instance.
Although it targets within-instance generalization, this approach can be used across instances.
Another approach, called \textit{extrapolation}, was recently explored~\cite{Pre22}. This method requires learning ground constraints for \textit{multiple} instances, which are then extrapolated to a new instance with different parameters.  
A limitation is that because it is based on genetic programming, 
it tends to learn contrived, non-interpretable expressions.
A third approach is to learn interpretable parameterized forms of constraints \textit{during} the acquisition process~\cite{LalEtc,kumar2022learning}, given examples from different instances of the problem, with the corresponding instance parameters also explicitly given.
In~\cite{LalEtc} an inductive logic programming approach is used to learn rules of the form $\mbox{\textit{condition}} \Rightarrow \mbox{\textit{constraint}}$. On the other hand, \textsc{Count-CP}~\cite{kumar2022learning} first learns instance-specific constraints, 
which are then grouped 
to obtain \textit{first-order constraints} that can generalize to unseen instances. 

In the literature, there is no standardized method for representing generalized constraint specifications. As a result, the different approaches mentioned target different generalizations based on pre-defined partitions of variables.
This prevents them from capturing the diverse range of constraint specifications that may exist in CP models. Finally, methods relying on symbolic search can be expected to struggle in the presence of noise in the input ground constraints, which can obscure the underlying patterns they search for.

In this paper, we first formalize the elements of a parameterized constraint specification, capturing the requirements of a constraint problem in a generic and parameterized way. Then, we propose \textsc{GenCon}, a novel approach to learn such specifications from the constraints of given instances, leveraging the capabilities of statistical machine learning (ML) to learn complex functions from labeled data.
In more detail, our contributions are:

\begin{itemize} 

\item We introduce an approach for generalizing from one or more given problem instances using constraint-level classification, and we present a parameterized feature representation to capture the constraint specifications.

\item For classifiers that allow decision rule extraction, we present a method to translate them into interpretable constraint specifications. These can then be used to generate ground constraints for new problem instances.

\item To enable the use of our method with classifiers that do not allow rule extraction, we present a generate-and-test approach that can be used with any classifier.

\item We conduct a comprehensive experimental evaluation of our methods, using different classifiers across a range of problem classes and instances. We also show that our approach is robust to noise in the given ground constraints.
\end{itemize}

\section{Background}
\label{sec:back}
We now introduce the necessary concepts used in the paper.

\paragraph*{Constraint Satisfaction Problems}

A \textit{constraint satisfaction problem} (\textit{CSP}) is a triple $P=(V, D, C)$, defining:

\begin{itemize}

\item a set of $n$ decision variables $V = \{v_1, v_2, ..., v_n\}$, representing the entities of the problem,

\item a set of $n$ domains $D = \{D_{v_1}, D_{v_2}, ..., D_{v_n}\}$, with $D_{v_i} \subset \mathbb{Z}$ being the domain of $v_i \in V$,

\item a constraint set $C = \{c_1, ..., c_t\}$. over the variables in $V$.

\end{itemize}

A \textit{constraint} $c$ is a pair ($\mathit{rel}(c)$, $\mathit{var}(c)$), where $\mathit{var}(c)$ $\subseteq V$ is the \textit{scope} of the constraint, and $\mathit{rel}(c)$ is a relation over the domains of the variables in $\mathit{var}(c)$, restricting their allowed value assignments. The \textit{arity} of the constraint, denoted as $|\mathit{var}(c)| = \text{arity}(\mathit{rel}(c))$, indicates the number of variables involved. 
The set of solutions of a constraint set $C$ is denoted by $sol(C)$. 

\paragraph*{Constraint Acquisition}
In Constraint Acquisition (CA), the pair $(V, D)$ is called the \textit{vocabulary} of the problem at hand and is common knowledge shared by the user and the system. Besides the vocabulary, the learner is also given a \textit{language} $\Gamma$ consisting of a broad range of {\em fixed arity} constraint relations that may exist in the problem at hand. Using the vocabulary $(V, D)$ and the constraint language $\Gamma$, the system generates the \textit{constraint bias} $B$, which is the set of all expressions that are candidate constraints for the problem. 

The (unknown) target constraint set $C_T$ is a constraint set such that for every example $e$ it holds that $e \in sol(C_T)$ iff 
$e$ is a solution to the problem the user has in mind. The goal of CA is to learn a constraint set $C_L$ that is equivalent to the unknown target constraint set $C_T$.

\paragraph*{Machine Learning Classification}

ML classification is a supervised learning task that involves learning a function over a given dataset. The dataset, denoted as $\mathbf{E}$, is a collection of $N$ training examples, $\mathbf{E} = \{(x_1, y_1), (x_2, y_2), ..., (x_N, y_N)\}$. Each example is a pair $(x_i, y_i)$, where $x_i$ is a feature vector from the input space $X$ and $y_i$ is the corresponding class label from the output space $Y$. The feature vector \( \mathbf{x}_i \) is composed of \( m \) features, \( \mathbf{x}_i = (\phi_{i1}, \phi_{i2}, \ldots, \phi_{im}) \), with each feature \( \phi_{ij} \) being a (quantifiable) property of example $i$. In the case of 
classification, $Y$ is a set of possible class labels. 
An ML classifier aims to learn a function $f_{\theta}: X \rightarrow Y$, using a set of learnable parameters $\theta$. These parameters are adjusted during training to minimize a loss function $L(f_{\theta}(x), y)$ measuring the error between the predicted and actual class labels. 

\paragraph*{Decision Rules}
With the rising importance of explainable AI (XAI) and interpretable ML, various approaches focus on extracting decision rules from ML models~\cite{gilpin2018explaining}. These approaches represent the function $f_{\theta}: X \rightarrow Y$ with if-then rules, denoted as a set $R = \{r_1, r_2, ..., r_k\}$. Each decision rule $r_i$ is a pair $(Q_i, y_i)$, with $Q_{i}$ being a set of conditions and $y_i$ a class label. Each condition $q_{ij} \in Q_i$ is a function $q_{ij}: X \rightarrow \{0, 1\}$ that maps an example $x$ to a binary value indicating whether the given example satisfies the condition. For a rule $r_i = (Q_i, y_i)$ to be satisfied, all of its conditions need to be satisfied, i.e., $q_{ij}(x) = 1 \mid \forall q_{ij} \in~Q_i$. 

\section{Problem Definition}

Constraint problems are often not thought of as a single CSP, but as a set of requirements, with the specific instantiation of the ground CSPs depending on the values of some input parameters $\mathcal{P}$. We illustrate this with the following example, which we will also use as a running example. 

\begin{example}
\label{ex:examtt}
Consider a simplified exam timetabling problem with $s$ semesters and $\mathit{n}$ courses per semester. The goal is to schedule the courses' exams over $d$ days, each having $t$ timeslots. The parameters of the problem are $\mathcal{P} = \{s, \mathit{n}, d, t\}$. The requirements are that all exams must be scheduled in different timeslots, while exams of courses from the same semester must be on different days.
Different values for the parameters will lead to different ground CSPs for each problem instance. The parameters $\mathit{n}$ and $s$ determine the variables $V$ of the problem, while $d$ and $t$ determine their domains $D$. The parameter values also determine the set of constraints $C$. The problem contains \verb|different_day| constraints, which are defined over partitions of variables for courses in the same semester, and the allowed assignments depend on the timeslots per day $t$.
\end{example}

\begin{defn}
\label{def:problem_specification}
A \emph{parameterized constraint problem} consists of a set of parameters $\mathcal{P} = \{ p_1, p_2, \ldots, p_q \}$, and a function mapping each parameter instantiation $\mathcal{P}_{A} = \{(p_1, u_{A1}), (p_2, u_{A2}), \ldots, (p_q, u_{Aq}) \}$ onto a ground CSP with $(V_{A}, D_{A}, C_{A})$. The resulting tuple $(\mathcal{P}_{A}, V_A, D_A, C_A)$ is a \emph{problem instance}.
\end{defn}

Most CA techniques are aimed at learning the constraint set $C_T$ of a \textit{single} ground CSP, from examples of solutions and non-solutions of that instance. However, in learning, one is often interested in generalizing beyond the instance used for learning, across other instances of the same underlying parameterized constraint problem. 

\begin{defn} 
\label{def:generalization}
Given one or more problem instances described by tuples $(\mathcal{P}_A, V_A, D_A, C_A)$, the objective of generalization is to construct a function $F$ such that, for any \textit{target} problem instance with a vocabulary $(V_T, D_T)$, defined by a unique set of parameter values $\mathcal{P}_T = \{ (p_1, u_{T1}), (p_2, u_{T2}), \ldots, (p_q, u_{Tq}) \}$, $F(\mathcal{P}_T, V_T, D_T)$ will return the corresponding set of constraints $C_T$. 
That is, the aim is to learn the function $F$ from the given ground CSPs, to be able to accurately
determine the set of constraints $C_T$ for any target instance with parameters $\mathcal{P}_T$.
\end{defn}

\section{Constraint Specifications}

Being able to generalize constraint models involves finding such a function $F$ (definition~\ref{def:generalization}). Typically, in constraint problems, such a function $F$ can be decomposed into several inner functions -- which we model as \textit{constraint specifications} (CSs) -- each corresponding to a specific requirement of the problem; for example "all courses must be scheduled in a different timeslot". The complete set of constraints $C_T$ of an instance $T$ is then the union of the sets of constraints produced by each inner function. 

Each requirement is modeled by a constraint specification, which defines how to derive the pairs ($\mathit{rel}(c)$, $\mathit{var}(c)$) for any target instance $T$, using the parameter values $P_T$ and the corresponding vocabulary $(V_T, D_T)$. We consider the following three key elements of a CS: 

\begin{enumerate}
    \item Constraint \textit{relation}. 
    The relation $\mathit{rel}(c)$ of each constraint in this CS, which may optionally depend on parameters that determine constant values involved in the relation.
    
    \item Variable \textit{partition(s)}. Typically, a pattern that appears in a constraint model concerns certain \textit{partitions} of variables of the problem and is applied to sequences of variables in this partition. Such partitions can be the dimensions (e.g., rows and columns) of the (multi-dimensional) matrix the variables are given in, or based on latent dimensions in this tensor. 
    
    \item \textit{Sequence conditions}. These define which scopes within a partition of variables to apply the constraints to. It is common to have a constraint apply to all possible scopes in a partition. This is done by using the sequence \verb|all_pairs| for binary constraints, or more generally, the sequence \verb|combinations|, to find \textit{combinations} of size $arity(r)$ (the arity of the given relation). However, there may also be \textit{sequence conditions}, restricting the variable combinations that should be taken as scopes. 
\end{enumerate}

Using these three key elements, we now formally define constraint specifications:

\begin{defn}
\label{def:spec}
A \emph{constraint specification} (CS) is a triple $(r, G, S)$, defining 
\begin{itemize}
    \item 
a \textit{relation} $r$, along with any parameters defining its constants, 
   
    \item 
a \textit{variable partition function} \( G: V_T \to \mathcal{P}(V_T) \), that partitions a given set of variables $V_T$ into subsets based on certain characteristics,
    
    \item 
a set of \textit{sequence conditions} $S$ restricting the scopes to which the constraints are applied.

\end{itemize}
\end{defn}

\noindent A CS can generate the corresponding ground constraints of an instance $T$ using the following generator template:

\begin{small}
\begin{Verbatim}[commandchars=\\\{\}]
Foreach Y \(\in \mathcal{P}(V_T)\):
  Foreach scope \(\in\) combinations(Y, arity(r),
                                \(S\)):
    c \(\leftarrow\) (rel(c) = r, var(c) = scope)
\end{Verbatim}
\end{small}

\begin{example}
Consider a problem with the requirement ``Values of consecutive variables in the same row must differ". The CS modeling this requirement uses the ``$\neq$" relation. The different partitions used in the CS are the rows of the variable matrix. The CS alsone needs to express the sequence condition that the variables need to be consecutive, as it does not apply to all pairs of variables in the same row.
In a modeling language, this requirement would be modeled as:
\begin{small}
\begin{Verbatim}[commandchars=\\\{\}]
Foreach row \(\in\) all_rows:
  Foreach var1, var2 \(\in\) consecutive_pairs(row): 
        c \(\leftarrow\) var1 \(\neq\) var2
\end{Verbatim}
\end{small}

This is equivalent to our formally defined CS generator, where $\mathcal{P}(V_T)$ corresponds to \verb|all_rows| and \verb|consecutive_pairs| corresponds to \verb|combinations| with arity 2, and sequence condition \verb|column(var1) - column(var2) == 1|. 
\end{example}

\hide{
\begin{example}
Recall the exam timetabling problem from Example~\ref{ex:examtt}. This problem has two requirements:
\begin{itemize}
    \item all courses must be scheduled in a different timeslot,
    \item exams of courses from the same semester must be scheduled on different days.
\end{itemize}
Let us focus on the CS modeling the requirement ''courses from the same semester cannot be scheduled the same day``.
This CS uses the \verb|different_day| relation.
The partitioning function groups variables within the same semester, which are commonly given in the same row in the variable matrix. It contains no sequence conditions, as the constraint applies to all pairs of courses in each semester. 

Note that in a modeling language this requirement is typically modeled as:
\begin{small}
\begin{Verbatim}[commandchars=\\\{\}]
Foreach row \(\in\) all_rows:
  Foreach scope \(\in\) all_pairs(row):
    c \(\leftarrow\) different_day(scope)
\end{Verbatim}
\end{small}

This is equivalent to our formally defined CS generator, where $\mathcal{P}(V_T)$ corresponds to \verb|all_rows| and \verb|all_pairs| corresponds to \verb|combinations| with arity 2, without any conditions. 

\end{example}
}

\section{Generalizing Constraint Models}
\label{sec:gencon}

To generalize beyond one or more known instances and learn the CSs of a problem, we propose an approach named \textsc{GenCon}. The key idea is to use ML to identify patterns in the constraints of the known instance(s) and reconstruct the CSs of the problem. This is especially promising because, recently, an approach using probabilistic classification during active CA~\cite{tsouros2023learning} demonstrated that ML classifiers can effectively detect patterns within the learned constraint network.

\begin{figure*}[tb]
\centering
    \includegraphics[width=\textwidth]{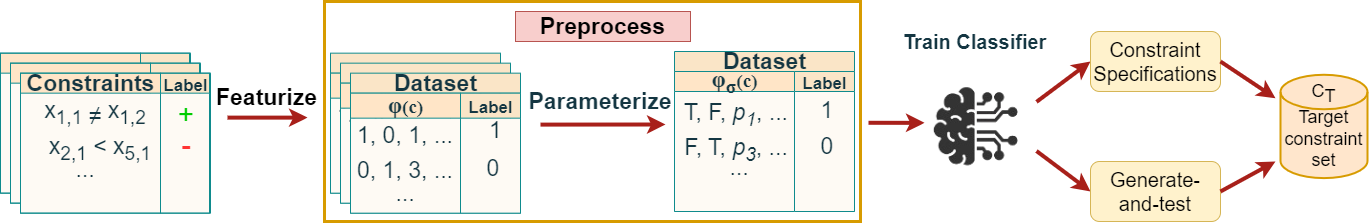}
     \caption{\textsc{GenCon}: Generalizing constraint models through constraint classification, using a parameterized feature representation of constraints}
 \label{fig:genacq-train}
\end{figure*}

\textsc{GenCon} is shown in Figure~\ref{fig:genacq-train}. 
The given set of ground constraints in the input instance(s) is used in order to \textit{train a classifier} to predict for any constraint whether it belongs to the set of constraints of any target instance of the problem. For this, we use a (parameterized) feature representation of the constraints, whose design is inspired by the different elements of CSs discussed in the previous section. 
For classes of classifiers allowing the extraction of decision rules, we directly translate these rules to the CSs of the problem, which can produce the ground constraints of any target instance. When decision rules cannot be extracted, a generate-and-test approach is used instead.

\subsection{Building the Dataset}

We build a dataset on the constraint level, i.e., its examples correspond to individual constraints. Given a set of constraints $C_A$ for a problem instance $A$, and a distinct set of constraints $C_{A}^-$, consisting of constraints that are not part of the model, we define a dataset $\mathbf{E}$, wherein each training example is represented as a tuple $(\mathbf{x}_i, y_i)$, corresponding to a constraint $c_i \in C$. For each example $(\mathbf{x}_i, y_i)$, we have $\mathbf{x}_i = \phi_{\sigma}(c_i,\mathcal{P})$, which is a parameterized feature representation of constraint $c_i$,  and $y_i = [c_i \in C_A]$, a Boolean label that indicates whether $c_i$ is part of the set of true constraints or not. 
\begin{equation}
\begin{split}
\mathbf{E} = \{ (\mathbf{x}_i, y_i) \mid \mathbf{x}_i = \phi_{\sigma}(c_i, \mathcal{P}) \land \text{ } y_i = [c_i \in C_A], \\
\forall \text{ } c_i \in \{ C_A \cup C_{A}^- \} \}, 
\end{split}
\end{equation}

However, realistically, we may only have access to the set of true constraints $C_A$ for each problem instance. 
But we also need a set of constraints $C_{A}^-$ consisting of constraints that will have a negative label, for the classifier to learn how to distinguish between the classes. To produce this set, we first generate a set of constraints $B_A$, using as a language $\Gamma$ all relations detected in the given set of constraints $C_{A}$, i.e, $\Gamma = \{ \mathit{rel}(c) \mid c \in C_A \}$. The bias $B_A$ is created by applying each relation in $\Gamma$ to all possible scopes in $V_A$. 
The set $C_{A}^-$ then consists of all constraints in $B_A$ that are not part of the given instance(s), i.e., $C_{A}^- = B_A \setminus C_A$.

\subsection{Parameterized Feature Representation}
In our approach, we propose a framework for the (parameterized) feature representation of constraints, targeted at learning patterns based on the different elements of CSs discussed above.
For any constraint $c$, the classifier expects a fixed-size feature representation $\phi(c)$ as input.
As shown in the middle of Figure~\ref{fig:genacq-train}, this feature representation $\phi(c)$ is then transformed to a parameterized version, denoted by $\phi_{\sigma}(c)$, to be able to learn patterns across instances with different parameter values.

\paragraph*{Feature Representation.}
The feature representation $\phi(c)$ must be designed based on the different elements of CSs that we want the classifier to detect in the problem, i.e., the relations, partitioning functions, and sequence conditions. Hence, it must contain features that describe the constraint relations, variable characteristics that can be used to recognize partitions of the variables, and other attributes that can play a role in the sequence conditions. Based on this, we construct a feature representation consisting of three groups of features:

\begin{enumerate}
    \item \textbf{Relation features}: Features that capture properties of the relation $rel(c)$ of a given constraint $c$, along with numerical values of the constants present in the constraint.
    \item \textbf{Partitioning features}: Features describing whether the variables in the scope of constraint $c$ have characteristics \textit{in common} that can be used in the partitioning function. These characteristics can be problem-specific variable properties (e.g., in what semester a course takes place in exam timetabling), or based on information regarding the structure the variables were given in. For example, in many cases, the variables $V$ are given in the form of a matrix or tensor, and the position of each variable in this tensor often plays a crucial role in the partitioning function of the CSs. 
    \item \textbf{Conditioning features}: These features describe how the variables in the scope of the constraint relate in different ways, to capture sequence conditions that may exist in the CSs of the problem. 
    For example, a constraint may only apply to pairs of variables that are a certain distance away from each other in the variable tensor. Thus, the distance between the variables in a constraint's scope may be included as a conditioning feature. Note that the partitioning features can also be used to capture the sequence conditions, since they describe whether the variables share a certain property or not. For example, a sequence condition may state that the variables must not be part of the same row in the variable matrix.
\end{enumerate}

The grammar of relations, partitioning functions, and sequence conditions used can be considered as the inductive bias of our method. The feature representation needs to be able to capture the CSs existing in the problem at hand. 
In our implementation, a proof-of-concept feature representation was used based on structural properties of the variables matrix, as matrix modeling is common and beneficial in CP~\cite{flener2001matrix}, with no problem-specific variable attributes.\footnote{More information regarding the feature representation used in our implementation can be found in the appendix.} 
In more detail:
\begin{enumerate}
    \item We used 3 relation features, describing the name of the constraint relation and its constant values.
    \item As candidate attributes for partitioning, we used the \textit{indices} of the variables in the dimensions of the tensor they were given in, along with \textit{latent dimensions} that may be discovered using the problem parameters. We include one partitioning feature for each (latent) dimension, expressing whether all variables in the constraint's scope share the same index in them.
    \item As additional conditioning features, we used the average difference between the variable indices in each dimension and latent dimension.
\end{enumerate}

\paragraph*{From numerical to categorical features over parameters.}

As the goal is to generalize beyond a single problem instance, the feature representation of the constraints should capture the characteristics of the constraints in a generic, parameterized way. Numerical attributes of the constraints typically are not static across instances but depend on parameters of the problem; e.g., in our running example, the constant present in the \verb|different_day| constraints depends on the timeslots-per-day parameter and is not a static value. We want the classifier to be able to capture that. In this step of our approach, we thus replace numerical features
with categorical features over the parameters.
We do so using a numerical-to-categorical parameter mapping function
$\sigma : \mathbb{R} \rightarrow  \{\text{``NaN''}\} \cup \mathcal{P_A}$ (where $P_A$ is the list of named parameters and their value), defined as follows: 
\begin{equation}
\sigma(v) = 
\begin{cases} 
p_i &  v = u_i \mid (p_i, u_i) \in \mathcal{P_A} \\
\text{``NaN''} & \text{otherwise.}
\end{cases}
\end{equation}

For any numerical feature value that corresponds to a parameter value of the problem instance, function $\sigma$ replaces the feature by the corresponding parameter's name. 

Note that, in a constraint model, it is sometimes not the value $u$ of parameter $p$ that comes up directly in the features. Instead, a trivial arithmetic adaptation of the parameter value may be used, e.g., $u-1$, $u+1$ or the multiplication of two parameter values. To capture this, we extend the set of parameters $\mathcal{P}$ with these adaptations, along with the common basic constants 0 and 1, as is also done in \textsc{Count-CP}~\cite{kumar2022learning}.
For the new categorical features, there are thus $|\mathcal{P}|+1$  categories (where $\mathcal{P}$ is the extended set of parameters): one for every parameter, plus a  ``NaN'' in case none of the parameters match the given value. Our categorical features will thus be able to represent the constraint in a parameterized way. Although arbitrary constants cannot be captured this way, we make the assumption that every constant present is related to the parameters of the problem.

Also note that, when parameterizing the feature representation of a constraint, a single numerical feature value might correspond to multiple parameter values. When this occurs, one example is included in the dataset for each possible matching. Although this could add noise to the dataset, due to examples with a wrong parameter replacing the numerical feature, it ensures that the correct feature representations will definitely be included.

\subsection{Extracting Constraint Specifications}

Decision rules continue to be popular due to their interpretability, with methods existing to extract rules from various classes of classifiers~\cite{barakat2010rule,iqbal2012rule}. In this process, the goal is to derive a set of rules $R = \{r_1, r_2, \ldots, r_k\}$ that represent the classification function. Each rule $r_i$ specifies some conditions $Q_i$ on a subset of features, and a class label $y_i$, such that $Q_i \Rightarrow y_i$. In this context, rules that lead to a positive classification define the conditions for a constraint to be part of the target problem. Thus, these conditions can be converted into the CSs of the problem.
We now propose a method for extracting the interpretable CSs of the problem from such a set of extracted decision rules. The extracted CSs can then be used to generate the constraints of any given target instance.

In this context, rules that lead to a positive classification define the conditions for a constraint to be part of the target problem. Thus, these conditions can be converted into the CSs of the problem. As a result, our approach only operates on such positive-classification rules, iterating over their conditions to identify the relation, partitioning function and sequence conditions of each CS.

Our method is shown in more detail in Algorithm~\ref{alg:extract_cs}. First, the rules leading to a positive classification are extracted in $R_{pos}$ (line 1). 
Then, for each rule $r \in R_{pos}$ (line 3), a CS is constructed. The elements of the CS are first initialized (lines 5-7), and then the algorithm iterates over the rule conditions in $Q$ to construct the CS (line 8) as follows:
\begin{enumerate}
    \item \textbf{Relation Extraction}: Identify conditions in $Q$ related to \textit{relation features} (RF) (lines 10-11). These conditions determine which relations from $\Gamma$ are used, and which constants are used in these relations, if any.
    
    \item \textbf{Partitioning Function} (lines 12-14): Identify the partitioning function of the CS using conditions in $Q$ involving \textit{partitioning features} (PF). Use conditions that require certain characteristics to be \textit{equal} in the constraint's variables, and thus can be used to partition the variables based on them. 

    \item \textbf{Sequence Conditions}  (lines 15-18): Sequence conditions can be derived from both partitioning features and sequence condition features (SF). More concretely, if a condition $Q$ involves a partitioning feature, and requires certain characteristics to \textit{not} be equal, then this requirement is added to the sequence condition (lines 15-16). If a condition in $Q$ involves a sequence condition feature, it is also added to the sequence conditions (lines 17-18).
    
\end{enumerate}

\begin{algorithm}[tb]
\caption{Extracting Constraint Specifications}\label{alg:extract_cs}
\begin{algorithmic}[1]

\Require $R$: a set of decision rules, $\Gamma$: a set of relations, $RF$: a set of relation features, $PF$: a set of partitioning features, $SF$: a set of sequence condition features
\Ensure $CS$: a set of constraint specifications

\State $R_{pos} \leftarrow \{ r \in R \mid y(r) = \text{True} \}$
\State $CS \leftarrow \emptyset$

\ForAll {$r \in R_{pos}$}
    \State $Q \leftarrow \text{conditions}(r)$
    \State $rel \leftarrow \Gamma$
    \State $partition \leftarrow \emptyset$
    \State $seq\_cond \leftarrow \emptyset$
    
    \ForAll {$q \in Q$}
        \State $f \leftarrow \text{feature}(q)$
        
        \If {$f \in RF$}
            \State $rel \leftarrow rel \cap \{ \gamma \in \Gamma \mid q \text{ is satisfied by } \gamma \}$
        \ElsIf {$f \in PF$}
            \If {$\text{value}(q) = \text{True}$}
                \State $partition \leftarrow partition \cup \{ q \}$
            \Else
                \State $seq\_cond \leftarrow seq\_cond \cup \{ q \}$
            \EndIf
        \ElsIf {$f \in SF$}
            \State $seq\_cond \leftarrow seq\_cond \cup \{ q \}$
        \EndIf
    \EndFor
    
    \ForAll {$\gamma \in rel$}
        \State $cs \leftarrow \text{create\_CS}(\gamma, partition, seq\_cond)$
        \State $CS \leftarrow CS \cup \{ cs \}$
    \EndFor
\EndFor

\State \Return $CS$

\end{algorithmic}
\end{algorithm}

Finally, if more than one relation is allowed by the rule's conditions, we create one CS for each, retaining the partitioning function and sequence conditions (lines 19-21).

\begin{example}
\label{ex:cs}

Recall the exam timetabling problem from Example~\ref{ex:examtt}, which has two requirements:
``all courses must be scheduled in different timeslots'' and 
``exams of courses from the same semester must be scheduled on different days''.

In Figure~\ref{fig:dec-tree}, we can see a decision tree learned to classify constraints in this problem, using our parameterized feature representation. Recall that parameter $t$ represents the timeslots per day. We can extract the following decision rules from this tree by following the paths from the root to the leaves:
{\small
\begin{verbatim}
r1: Relation == "different_day" 
& Dim0_same == "false"
then 0 
r2: Relation == "different_day" 
& Dim0_same == "true" 
& Constant_parameter != "t" 
then 0 
r3: Relation == "different_day" 
& Dim0_same == "true" 
& Constant_parameter == "t" 
then 1 
r4: Relation == "!=" then 1 
\end{verbatim}
}

Rules $r3, r4$ are the positive-classification rules, which will be used to construct our CSs. The two CSs that will be extracted, along with their generators, are:
\begin{enumerate}
    \item CS$_1$: Relation: ``different\_day(t)'', Partitioning attribute: $dim_0\_same$, Sequence conditions: $\emptyset$
    \begin{small}
\begin{Verbatim}[commandchars=\\\{\}]
Foreach row \(\in\) all_rows:
  Foreach scope \(\in\) all_pairs(row):
    c \(\leftarrow\) (rel(c) = "different_day(t)", 
                    var(c) = scope)
\end{Verbatim}
\end{small}
    \item CS$_2$: Relation: ``!='', Partitioning attribute(s): None, Sequence conditions: $\emptyset$
    \begin{small}
\begin{Verbatim}[commandchars=\\\{\}]
  Foreach scope \(\in\) all_pairs(V):
    c \(\leftarrow\) (rel(c) = "!=", var(c) = scope)
\end{Verbatim}
\end{small}
\end{enumerate}

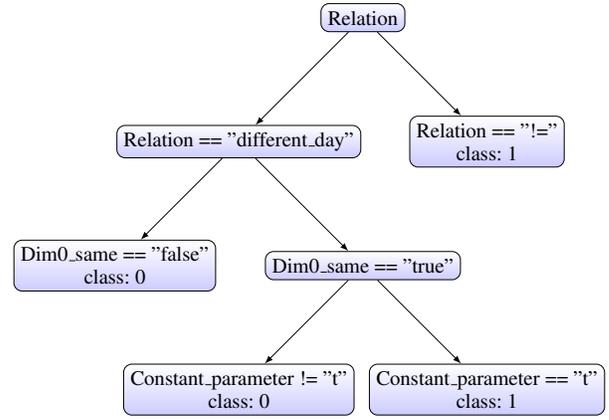
\begin{figure}[tb]
    \centering
      \resizebox{0.45\textwidth}{!}{
        \begin{tikzpicture}[
          edge from parent/.style={draw, -latex},
          sibling distance=12em,
          level distance=6em,
          every node/.style={shape=rectangle, rounded corners,
            draw, align=center, top color=white, bottom color=blue!20}
          ]
          \node {Relation}
            child { node {Relation == "different\_day"}
              child { node {Dim0\_same == "false"\\class: 0} }
              child { node {Dim0\_same == "true"}
                child { node {Constant\_parameter != "t"\\class: 0} }
                child { node {Constant\_parameter == "t"\\class: 1} }
              }
            }
            child { node {Relation == "!="\\class: 1} };
        \end{tikzpicture}
        }
    \caption{Decision Tree for Exam Timetabling in Example~\ref{ex:cs}}
    \label{fig:dec-tree}
\end{figure}
\end{example}

\subsection{Generate-and-Test}

To enable the use of our method even when decision rules cannot be extracted, we now present a generate-and-test approach that can be used with any classifier, as an alternative to extracting the CSs from interpretable classifiers. 

The intuition of this approach is the following:
Even if we cannot extract the CSs from the learned classifier $f_{\theta}$ explicitly, we know that it implicitly represents them. Thus, we can use the classifier itself to recognize the true constraints for any problem instance. Our \textit{generate-and-test} approach does so by generating a set of candidate constraints $B_T$ for the target problem $T$, using the language $\Gamma$ as described above. For relations with constants, the set $\mathcal{P}$ provides candidate values. To decide which of the constraints from $B_T$ to use, each of them is featurized, and the function $f_{\theta}$ predicts whether it should be part of the model. We keep all constraints with positive classification:
\begin{equation}
C_T = \{c \mid c \in B_T \land f_{\theta}(\phi_{\sigma}(c,\mathcal{P})) = \text{True} \}
\end{equation}

\section{Experimental Evaluation}

We now experimentally evaluate \textsc{GenCon}, using ground CSPs of different instances on a variety of benchmarks. We evaluate our approach both when the given sets of constraints are correct and when noise exists.
Noisy CSPs can result when the ground CSPs were themselves acquired using passive CA, on a noisy dataset of solutions and non-solutions, or on a dataset containing too few examples. We recognize two different types of noise in our setting:
\begin{enumerate}
    \item 
False positive (FP) noise, where the input set of constraints is not sound, also including wrong constraints. 
    \item 
False negative (FN) noise, where the input set of constraints is not complete, missing some true constraints.
\end{enumerate}

We aim to answer the following experiment questions:
\begin{itemize}

\setlength{\itemindent}{1.5em}

    \item [(Q1)] To what extent does \textsc{GenCon} effectively generalize ground CSPs?
    
    \item [(Q2)] What is \textsc{GenCon}'s performance when the input set of constraints also includes wrong constraints.

    \item [(Q3)] What is \textsc{GenCon}'s performance when the input set of constraints does not include all true constraints.

\end{itemize}

\subsection{Experimental Setup}

\textbf{Benchmarks}. 
We focused on using benchmarks that have different constraint specifications so that our method is evaluated in distinct cases. Namely, we used the following benchmarks that are commonly used in CA: Sudoku, Golomb, Exam Timetabling (ET) and Nurse Rostering (NR). In each benchmark, we used 10 instances with different parameters.\footref{myfootnote} We employed a challenging variant of leave-one-out cross-validation, referred to as \textit{leave-one-in cross-validation}: for each fold, we used just a single instance for training and the remaining nine instances for testing. We present the average results of this process.

\textbf{Metrics}. We evaluate each method by identifying the correctly generated constraints and the number of constraints missing from the model of the target instance(s). 
Based on that, we define as True Positives (TP) the correctly identified constraints, as False Positives (FP) the incorrectly identified constraints, and as False Negatives (FN) the missing constraints. 
Using the defined concepts, our evaluation is based on the following metrics that are common in ML:

\begin{itemize}
    \item 

\textbf{Precision (Pr)}: It measures the accuracy of the identified constraints in the target instance. 
A high precision score signifies a low rate of false positives.
When the precision score is 100\%, the predicted set of constraints is sound. 

    \item 

\textbf{Recall (Re)}: It measures the method's ability to identify all relevant constraints.
A high recall score indicates a low rate of false negatives.
When the recall score is 100\%, the predicted set of constraints is complete. 
\end{itemize}

\textbf{Comparison}. 
To obtain the CSs of the problem from extracted decision rules, we used Decision trees (DT) and the rule-based classifier CN2. Then, these CSs were used to generate the ground CSPs of the target instances. We also evaluated the generate-and-test approach with a variety of classifiers: Random Forests (RF), Naive Bayes (NB), Multi-layer Perceptron (MLP), and K-Nearest Neighbours (KNN). We used CN2, DT, RF, and NB with their default parameters and tuned the most important hyperparameters for MLP and KNN\footnote{Details can be found in the appendix.\label{myfootnote}}. We compare our method with the generalization approach used in \textsc{Count-CP}~\cite{kumar2022learning}. 

For the experiments that evaluate the impact of noise, we changed the ground CSPs for the input instances, injecting noisy constraints in them. We evaluated our method on 4 different levels of noise ($5\%, 10\%, 15\%, 20\%$) w.r.t. the original size of the input set of constraints $C_A$. To inject FP noise, we randomly add the respective percentage of constraints from the set constraints $C_{A}^-$ in $C_A$, 
while for FN noise, we directly remove constraints randomly from $C_A$.

\textbf{Implementation and hardware}
All experiments were conducted on a system with an Intel(R) Core(TM) i7-2600 CPU, 3.40GHz clock speed, with 16 GB of RAM.  All methods and benchmarks were implemented in Python. We used the CPMpy library~\cite{guns2019increasing} for constraint modeling, and the Scikit-Learn library~\cite{scikitlearn} for the ML classifiers, except CN2, for which the Orange library~\cite{demvsar2013orange} was used. For \textsc{Count-CP}, in the available implementation\footnote{https://github.com/ML-KULeuven/\textsc{Count-CP}} the generalization is mixed with learning, so we re-implemented it stand-alone.

\subsection{Results}

\begin{figure}[t!]
\centering
     \begin{subfigure}[b]{0.48\textwidth}
        \captionsetup{justification=centering}
         \centering
         \includegraphics[width=\textwidth]{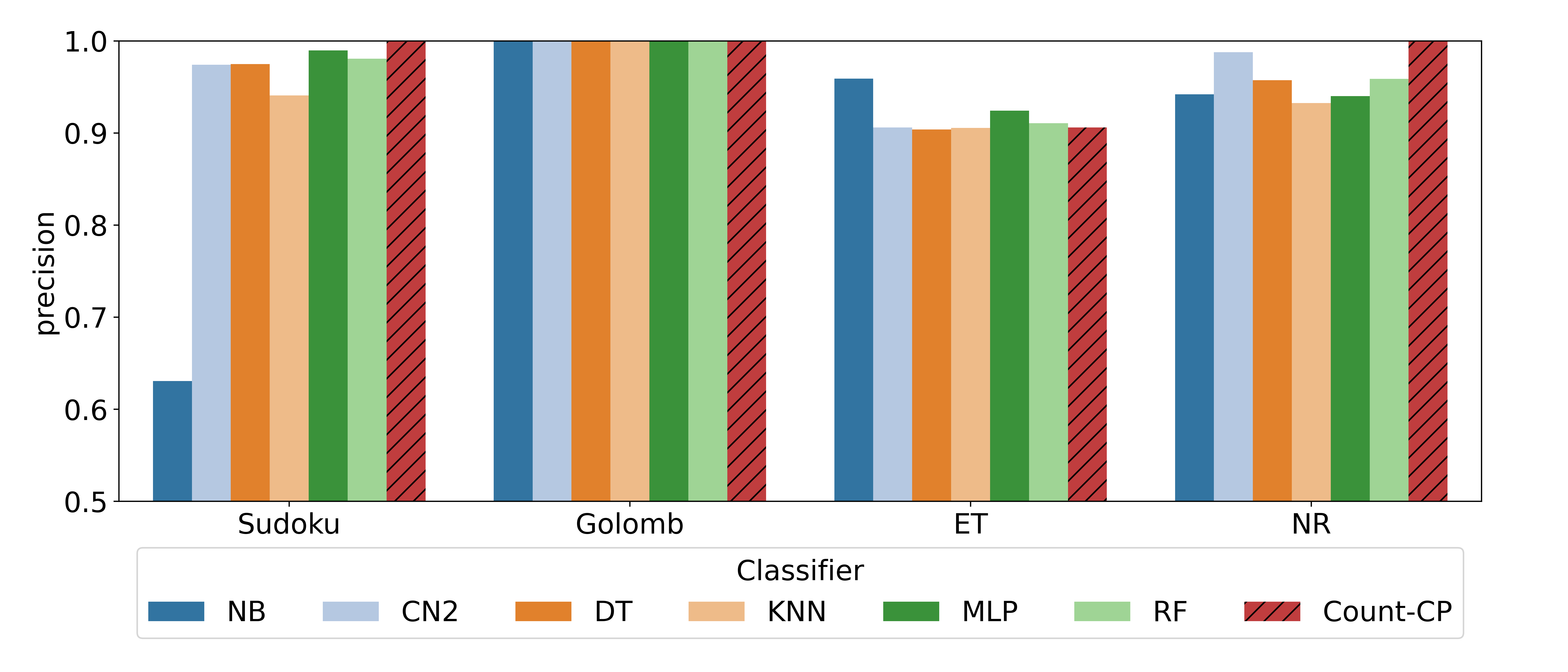}
         \caption{Precision}
     \end{subfigure}
    \begin{subfigure}[b]{0.48\textwidth}
        \captionsetup{justification=centering}
         \centering
         \includegraphics[width=\textwidth]{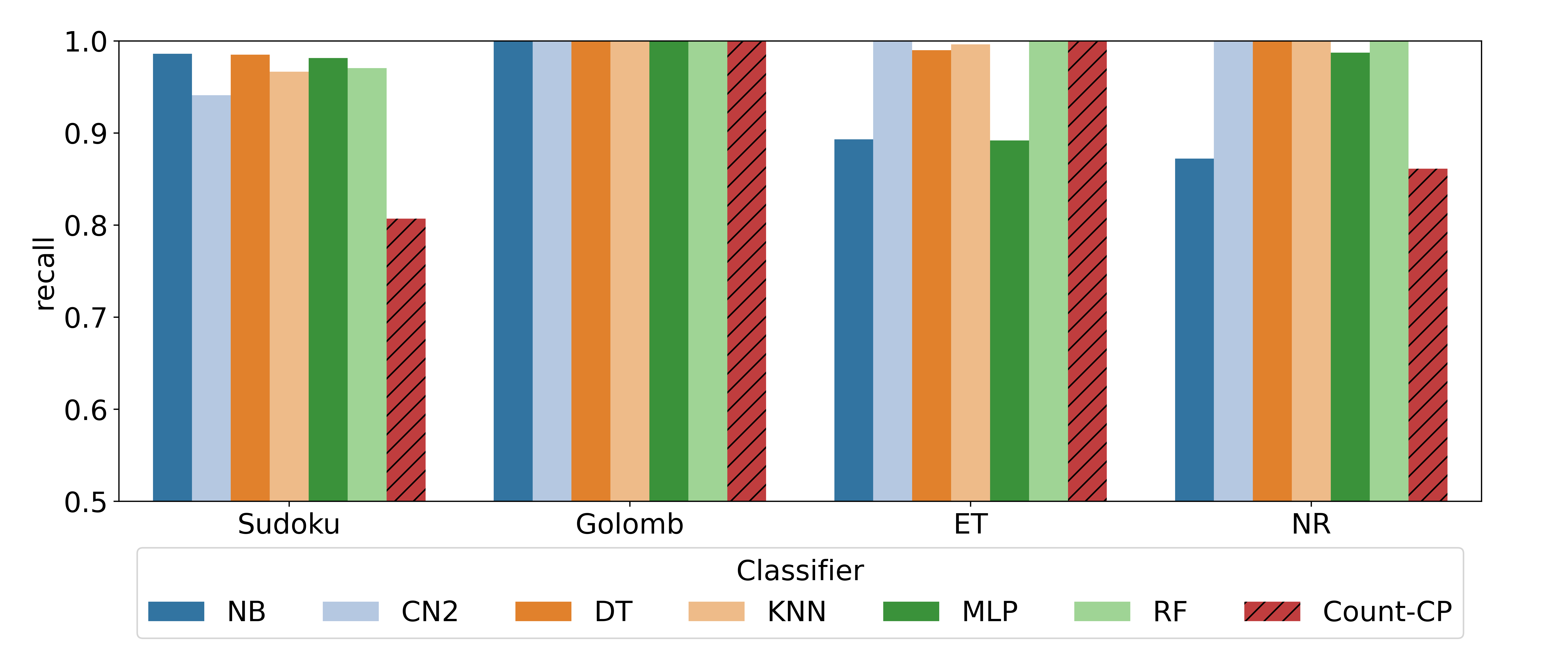}
         \caption{Recall}
     \end{subfigure}      
	\caption{Results comparing our method (using different classifiers) with \textsc{Count-CP} generalization}
 \label{fig:q1}
\end{figure}

\begin{figure*}[t!]
    \centering
    \begin{subfigure}[b]{0.22\textwidth}
        \centering
        \includegraphics[width=\textwidth]{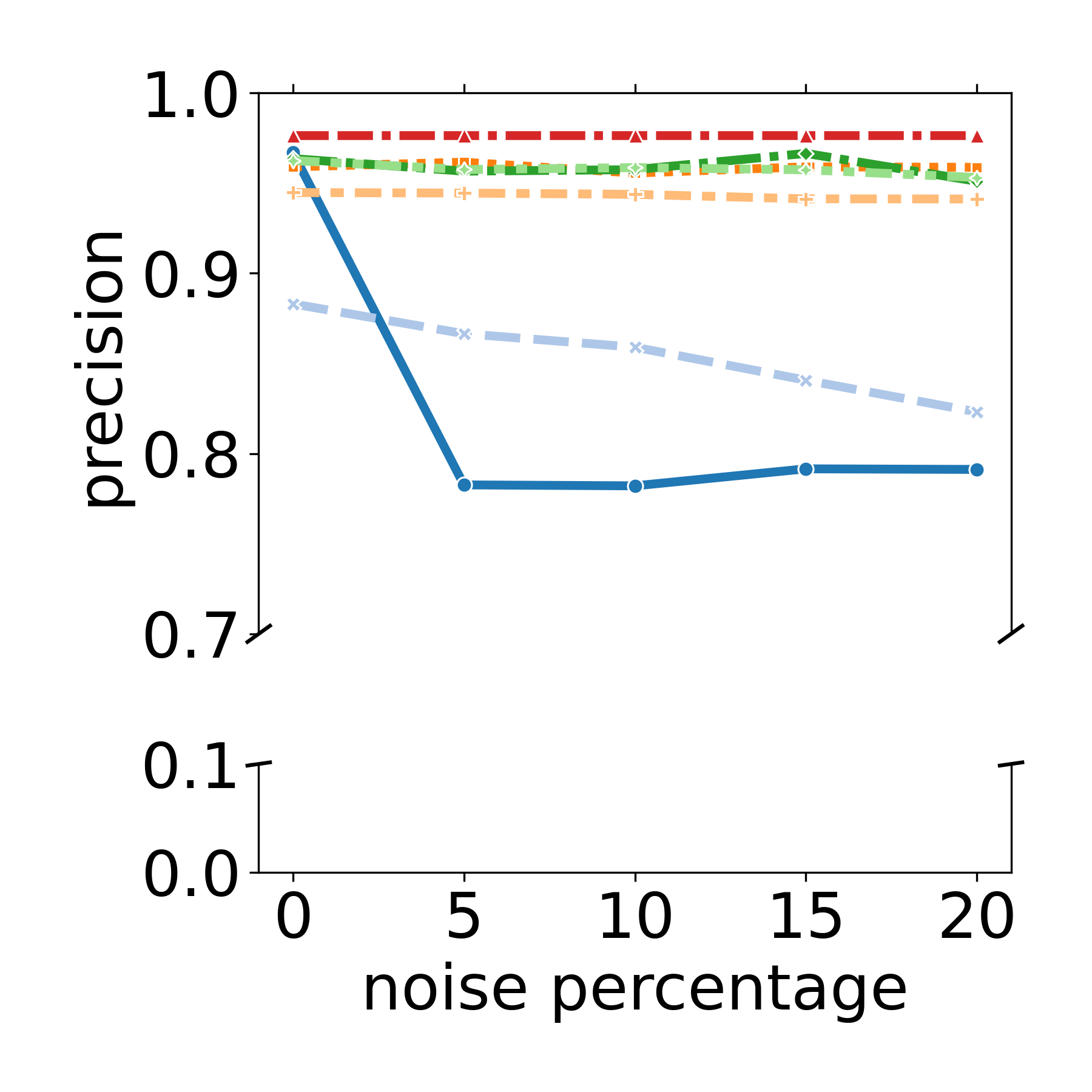}
        \caption{Precision with FP noise}
        \label{fig:fig1a}
    \end{subfigure}
    \hfill
    \begin{subfigure}[b]{0.22\textwidth}
        \centering
        \includegraphics[width=\textwidth]{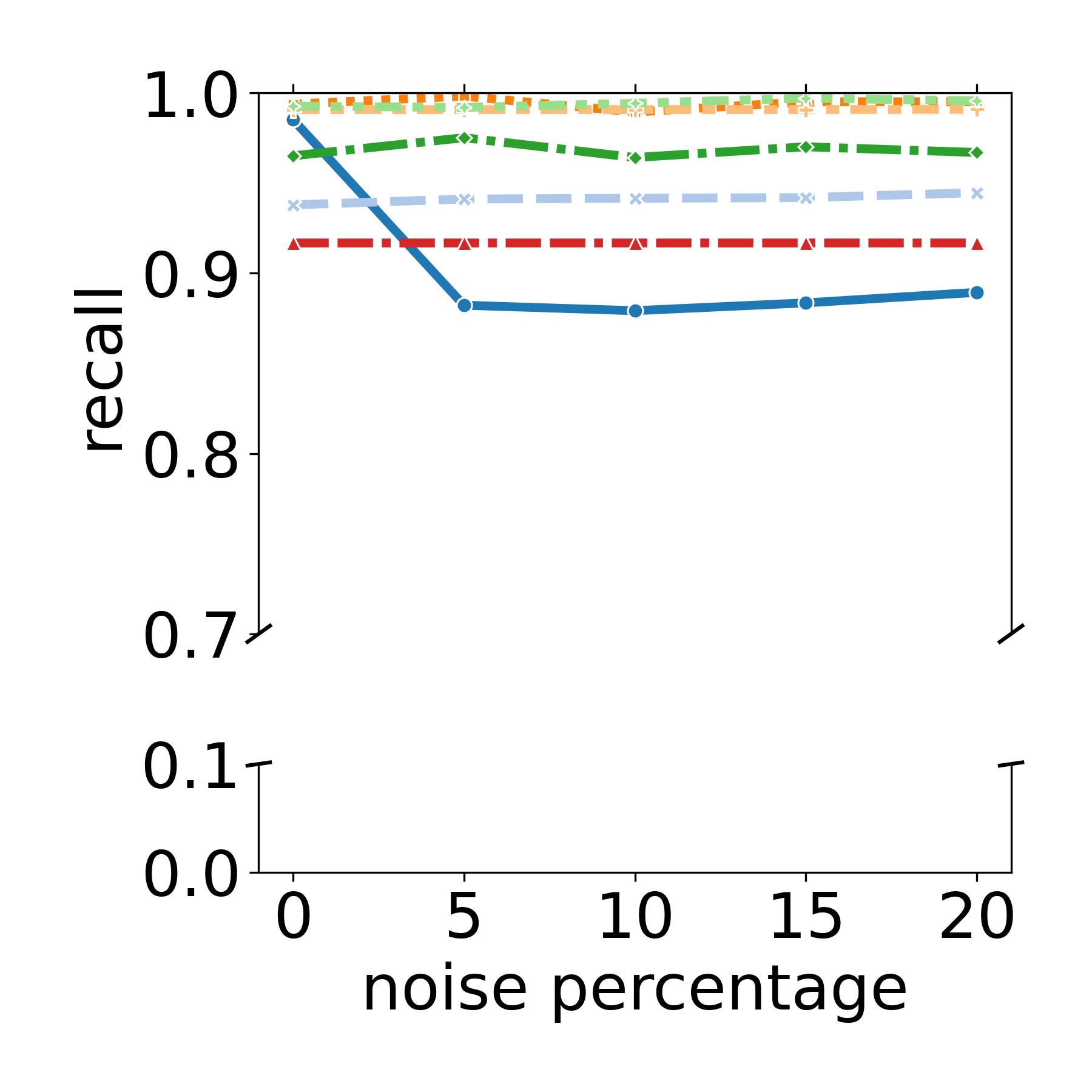}
        \caption{Recall with FP noise}
        \label{fig:fig1b}
    \end{subfigure}
    \hfill
    \begin{subfigure}[b]{0.22\textwidth}
        \centering
        \includegraphics[width=\textwidth]{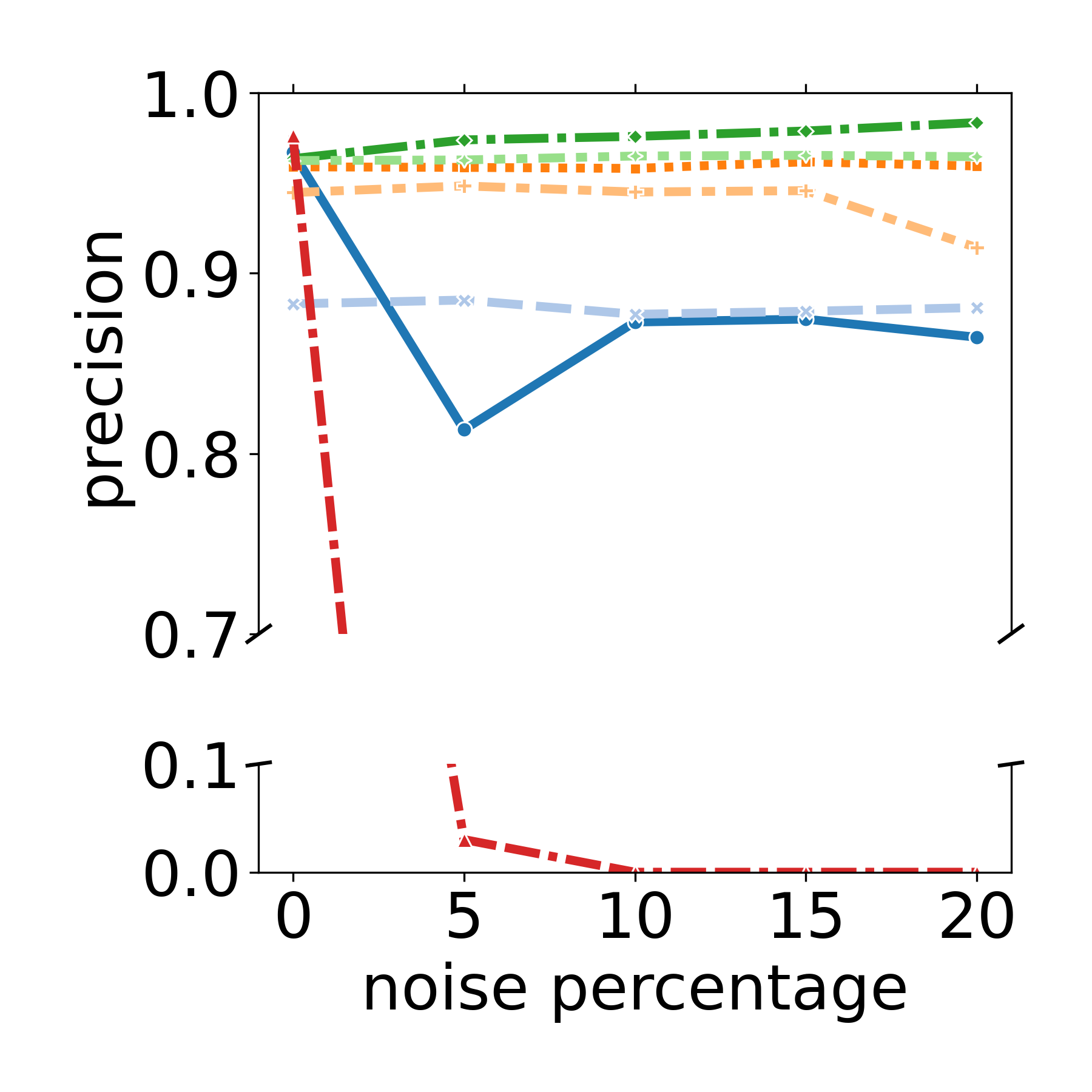}
        \caption{Precision with FN noise}
        \label{fig:fig2a}
    \end{subfigure}
    \hfill
    \begin{subfigure}[b]{0.22\textwidth}
        \centering
        \includegraphics[width=\textwidth]{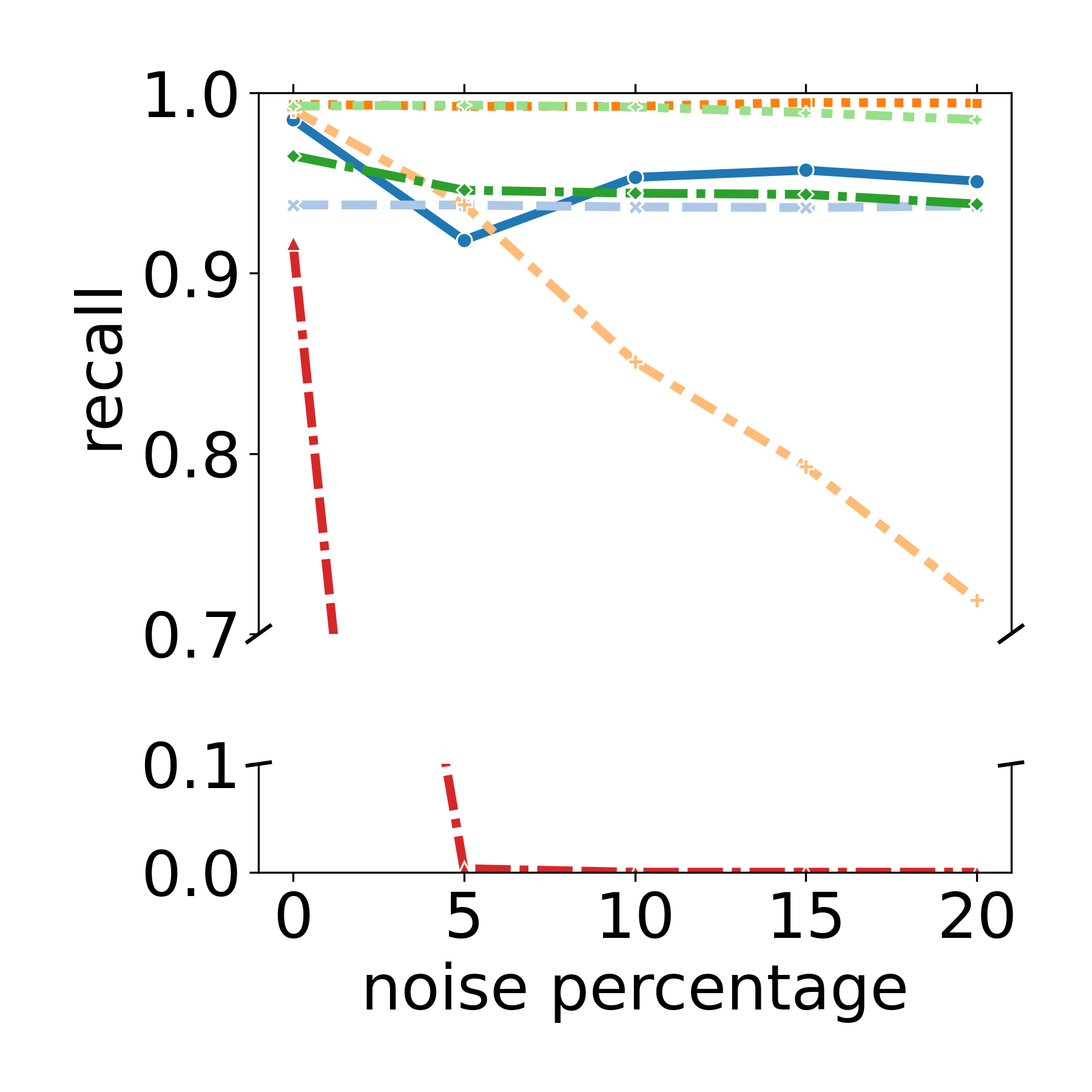}
        \caption{Recall with FN noise}
        \label{fig:fig2b}
    \end{subfigure}
    \begin{subfigure}[b]{0.10\textwidth}
        \centering
        \includegraphics[width=\textwidth]{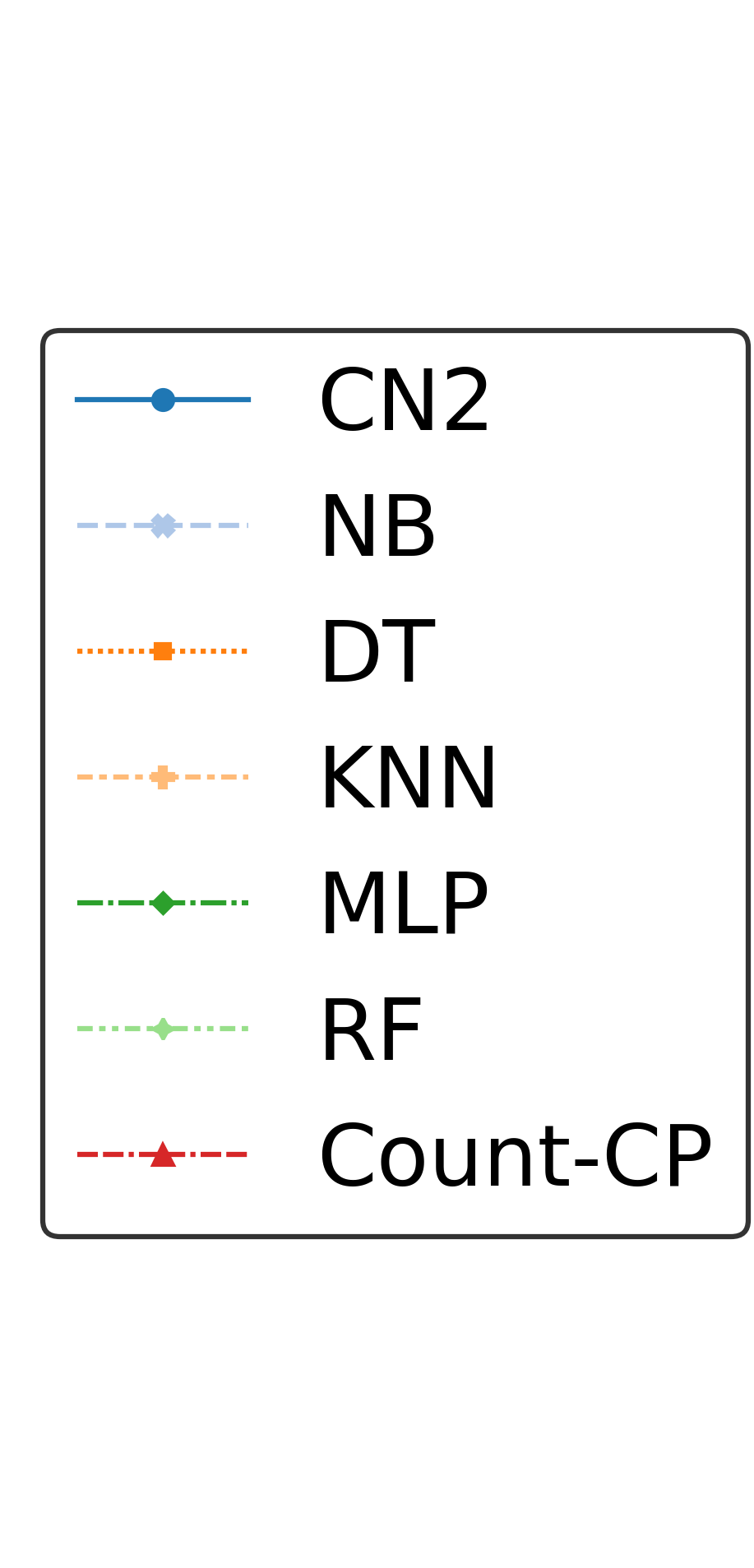}
        \label{fig:legend}
    \end{subfigure}
    \caption{(a) and (b) Results with the presence of FP noise in the input constraint model, (c) and (d) Results with the presence of FN noise in the input constraint model (best viewed in color).}
    \label{fig:combined}
\end{figure*}

\paragraph*{Q1: To what extent does \textsc{GenCon} effectively generalize ground CSPs?}

Figure~\ref{fig:q1} shows the results of \textsc{GenCon} using different classifiers, and of \textsc{Count-CP}'s generalization. Both the extraction of CSs for DT and CN2, as well as generate-and-test for the other classifiers, 
achieve high precision and recall in all benchmarks. The only exception is NB, which gets lower recall in ET and NR, and significantly lower precision in Sudoku. We believe that this is due to its feature independency assumption, which makes it hard for the classifier to recognise the relationship of the different features in the CSs. The benchmark that turned out to be the most difficult for all methods was ET, with the difficulty being recognizing the parameter of the CS regarding the \verb|different_day| constraints. The problem occurred in instances with many parameters with the same value, with all of them being recognized as part of the \verb|different_day| CS. This led to the generation of additional constraints in the target instances, lowering the precision. 

\textsc{Count-CP} also demonstrates good performance in general. Besides ET, where it presents the same issue as our method, it achieves $100\%$ \textit{precision} in the other three benchmarks. However, its main drawback is illustrated in the \textit{recall} results, as it was not able to capture all CSs in NR and Sudoku. In NR, the CSs regarding consecutive shifts cannot be captured, as \textsc{Count-CP} does not include sequence conditions in its generalization approach, and only searches for patterns that apply in all sequences of predefined partitions. In Sudoku, the block partitions are not automatically found.

\textsc{Count-CP} includes an option to manually give custom partitions as input, and in this case, its recall in Sudoku is increased to $100\%$. Notably, when we include special features for these custom partitions in our approach, the results with all classifiers also increase to $100\%$.

\paragraph*{Q2: What is the impact of FP noise in the performance of \textsc{GenCon}?}

Due to space limitations, we present the average results over all benchmarks for each method.\footnote{Detailed results per benchmark can be found in the appendix.}
The precision results are shown in Figure~\ref{fig:fig1a}, while the recall results are shown in Figure~\ref{fig:fig1b}.

We can observe that the CN2 classifier (and thus also the CSs extracted from it) and NB are the most sensitive to false positives, presenting increasingly worse precision scores when noise increases. This is because the learning approach of CN2 is not very tolerant to this kind of noise, overfitting in many cases to non-existing patterns. When any of the other classifiers are used in \textsc{GenCon}, their performance remains about as good as in the noiseless setting. The classifiers that already presented high scores in Q1 stay around 95-100\%, even when the noise percentage reaches 20\%.
Similarly, the \textsc{Count-CP} generalization approach keeps the same performance as in the original results without noise. That is because it only searches for specific partition patterns and symbolic expression bounds, and thus the randomly inserted constraints are directly disregarded.

\hide{
\begin{figure*}[t]
\centering
     \begin{subfigure}[b]{0.4\textwidth}
        \captionsetup{justification=centering}
         \centering
         \includegraphics[width=\textwidth]{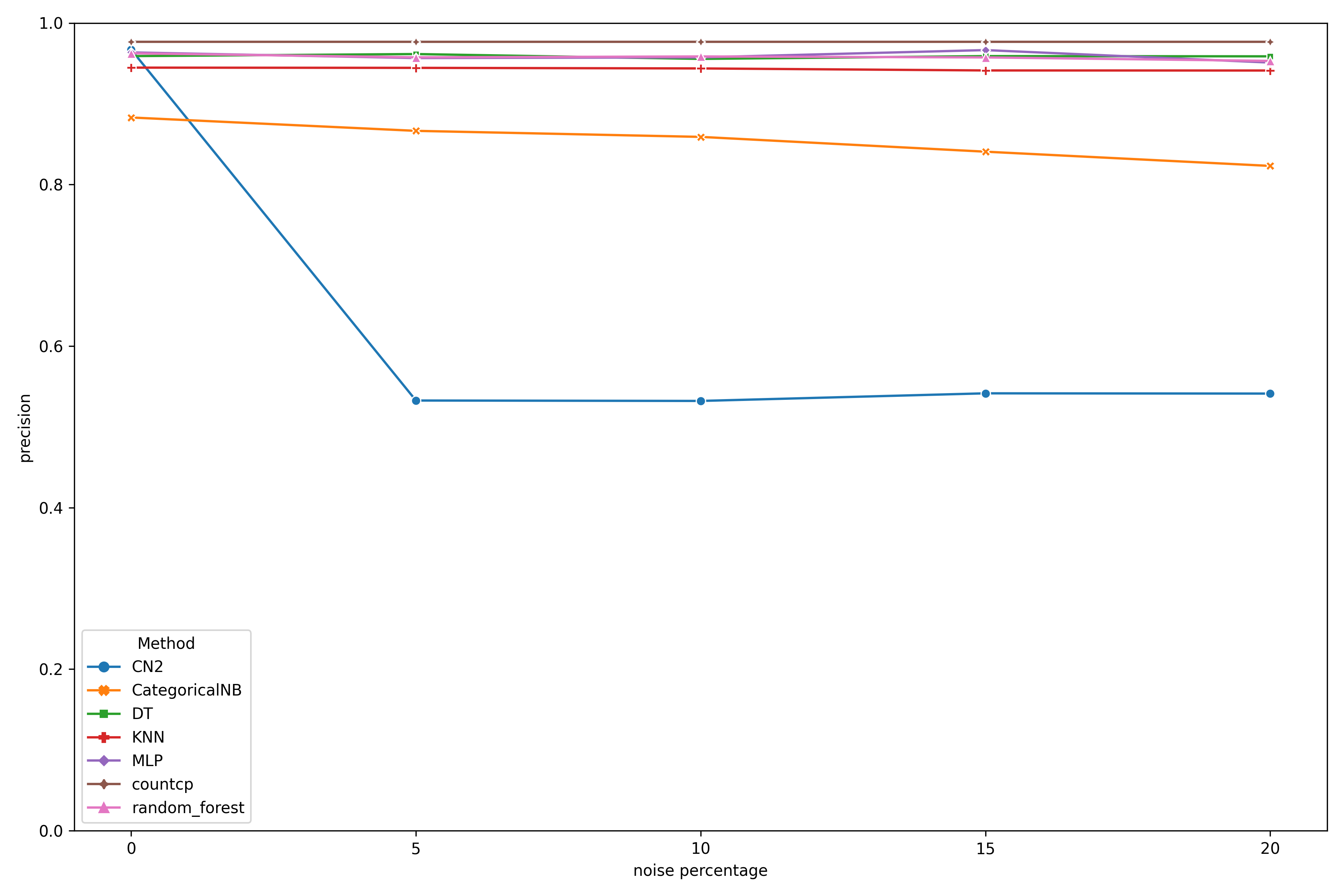}
         \caption{Precision}
     \end{subfigure}
    \begin{subfigure}[b]{0.4\textwidth}
        \captionsetup{justification=centering}
         \centering
         \includegraphics[width=\textwidth]{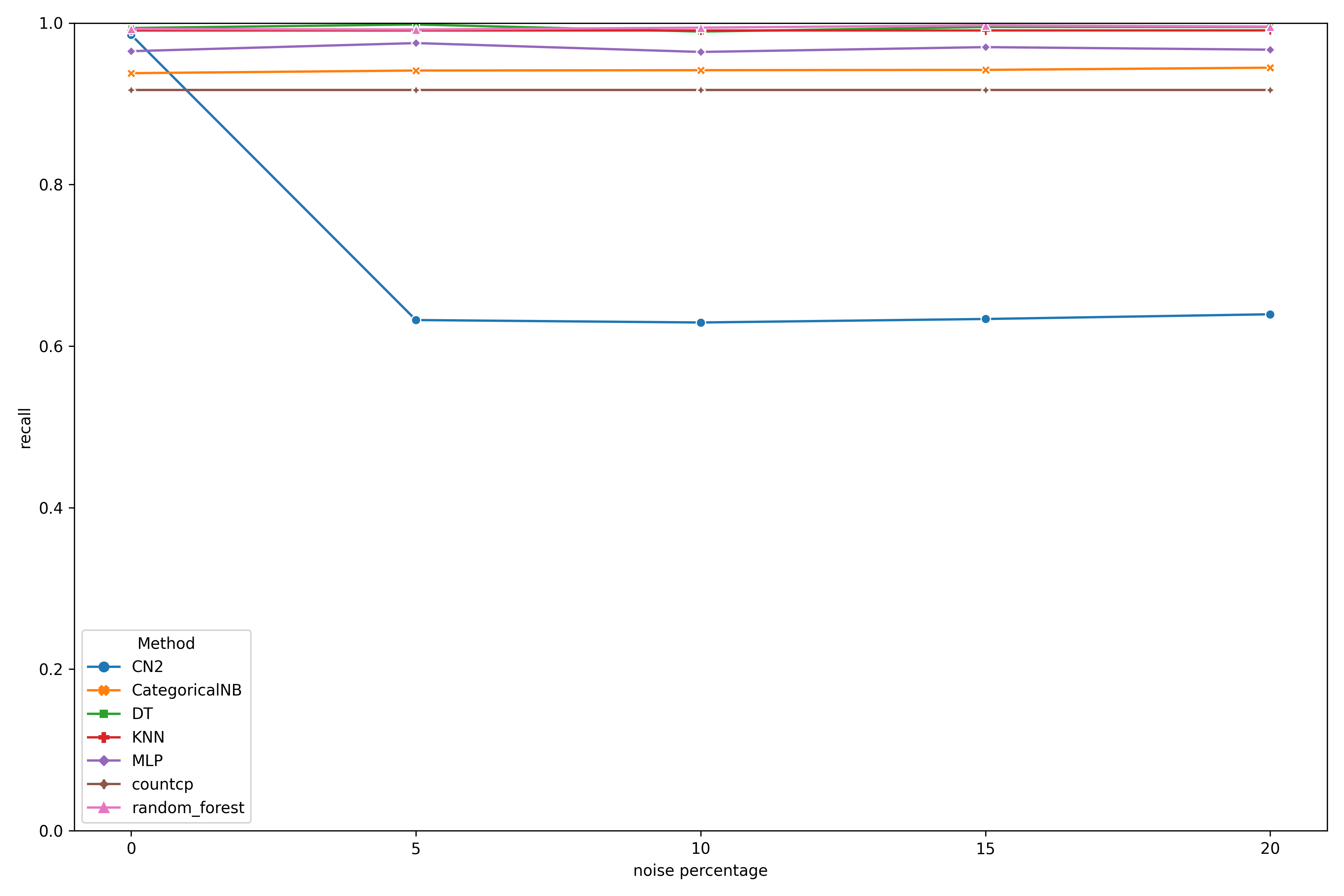}
         \caption{Recall}
     \end{subfigure}      
	\caption{Results with the presence of FP noise in the input constraint model, i.e., having additional (wrong) constraints in the input ground CSPs}
 \label{fig:fp_noise}
\end{figure*}
}

\paragraph*{Q3: What is the impact of FN noise on the performance of \textsc{GenCon}?}

As in Q2, we present the average results over all benchmarks. The precision results are shown in Figure~\ref{fig:fig2a}, while the recall results are shown in Figure~\ref{fig:fig2b}.

\hide{
\begin{figure*}[t]
\centering
     \begin{subfigure}[b]{0.4\textwidth}
        \captionsetup{justification=centering}
         \centering
         \includegraphics[width=\textwidth]{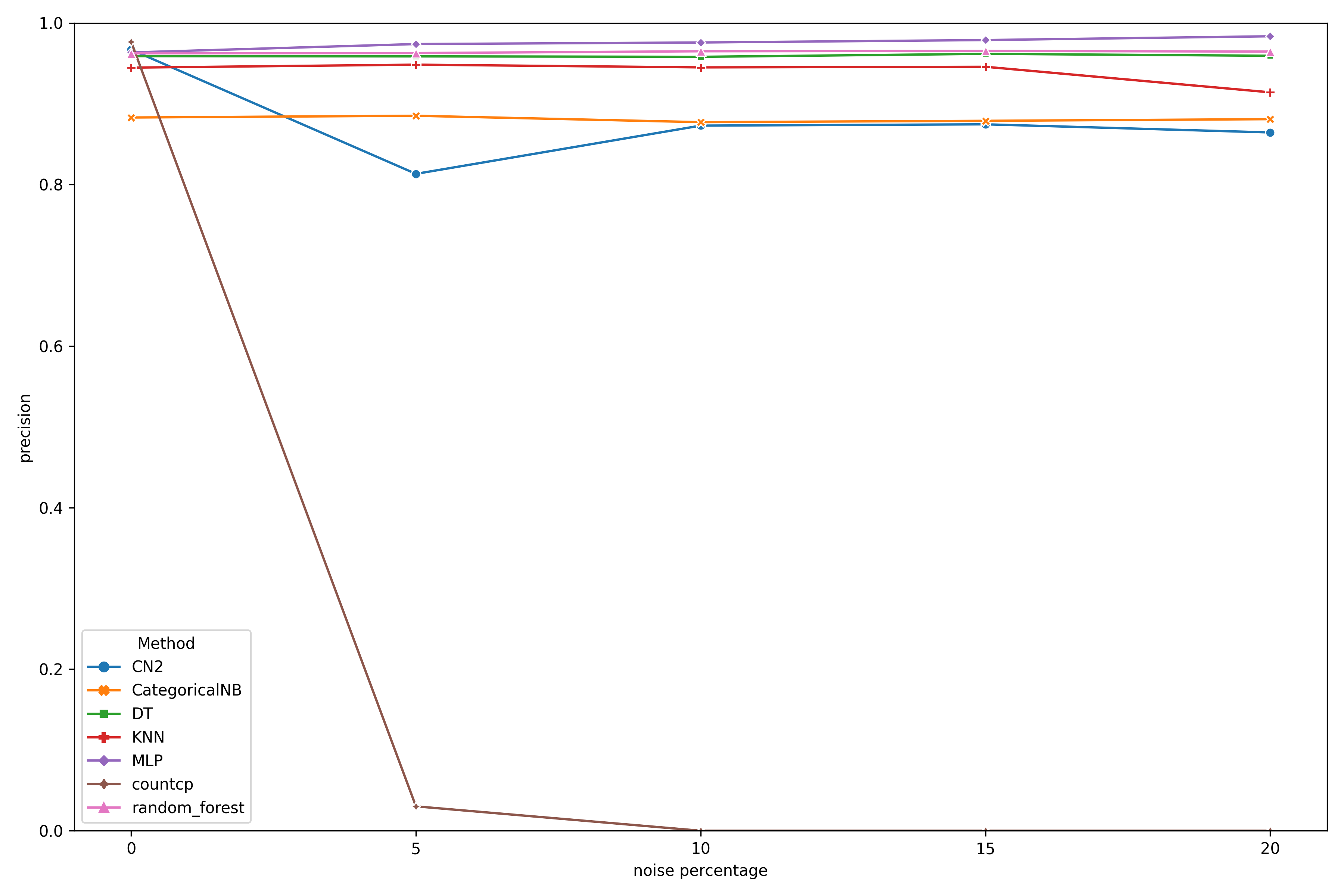}
         \caption{Precision}
     \end{subfigure}
    \begin{subfigure}[b]{0.4\textwidth}
        \captionsetup{justification=centering}
         \centering
         \includegraphics[width=\textwidth]{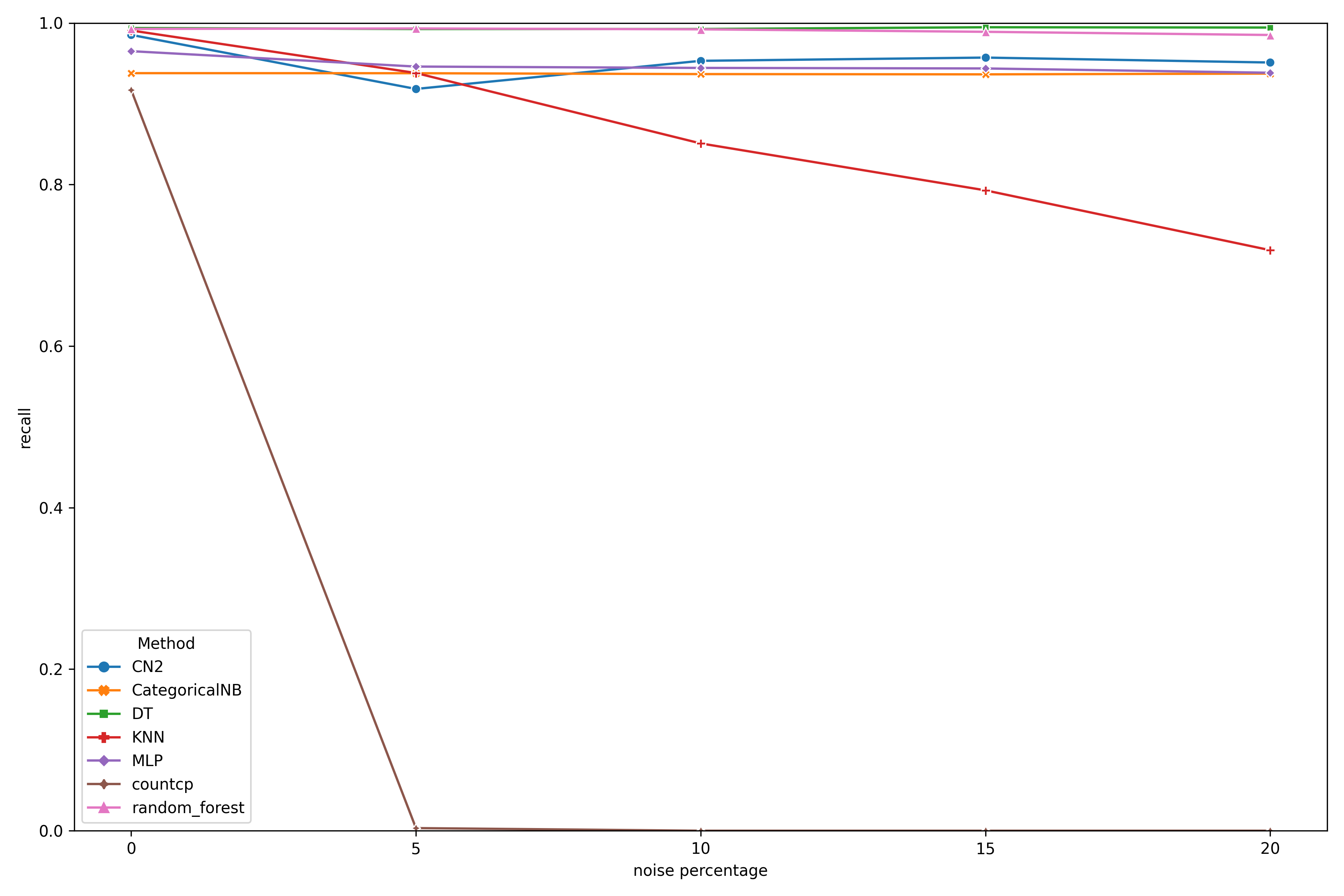}
         \caption{Recall}
     \end{subfigure}      
	\caption{Results with the presence of FN noise in the input constraint model, with correct constraints missing in the input ground CSPs}
 \label{fig:fn_noise}
\end{figure*}
}

We can observe that KNN is the most sensitive to FN noise, with worsening recall when more noise is added, meaning that it struggles to find all the constraints of the target instance. For CN2, the lack of noise tolerance shows up again, as in Q2, though precision and recall stay above $80\%$.
When any of the other classifiers is used, results remain good, even for up to $20\%$ noise. These results demonstrate the ability of our classification-based approach to generalize even in the presence of high percentages of noise.

On the other hand, in the presence of false negatives, the \textsc{Count-CP} generalization fails to detect any patterns and does not find any constraints in the target instances, as it searches for partitions in which all sequences of variables share a given constraint. Importantly, \textsc{Count-CP} fails to generalize, even when only 5\% noise is added.

\section{Conclusions}

CP models are typically defined by parameterized specifications, rather than a flat list of ground constraints. 
However, most CA methods focus on learning a single ground CSP for a specific instance. 
Our work addresses this limitation by generalizing ground CSPs to parameterized models using a constraint-level classification approach named \textsc{GenCon}. We showed how interpretable CSs can be derived from decision rules, and introduced a generate-and-test method for non-interpretable classifiers.
Our evaluation indicates that \textsc{GenCon} achieves high accuracy and robustness, even for high levels of noise, highlighting the potential of ML-based techniques for generalizing constraint models and making CA more robust. 
We recommend using decision trees as the classifier of choice, as they facilitate the extraction of interpretable CSs while presenting strong performance.

Promising avenues for future work include exploring active learning to enhance generalization; and using generalization during interactive constraint learning to reduce queries, leveraging also the noise robustness demonstrated here. Additionally, \textsc{GenCon} can be applied during passive CA, enabling the learning of constraint models from a limited amount of solutions and non-solutions across various instances, a scenario common in real-world applications.

\newpage

\section{Acknowledgments}
This research received funding from the European Research Council (ERC) under the EU Horizon 2020 research and innovation programme (Grant No. 101002802, CHAT-Opt); the Science Foundation
Ireland under Grant No. 12/RC/2289-P2 which is co-funded under the
European Regional Development Fund; and the Research Foundation-Flanders (FWO) (Grant No. 11PQ024N).

\bibliography{paper}

\newpage
~\newpage

\appendix

\section{Appendix}
\subsection{Feature representation used}

An overview of the feature representation we used in our implementation is given in Table~\ref{tab:features}. 

\begin{table}[ht]
    \centering
    \caption{Feature representation of constraints}
    \label{tab:features}
    \begin{tabular}{|p{0.015\textwidth}|p{0.25\textwidth}|}
        \hline
        ID & Name \\
        \hline
        \multicolumn{2}{|c|}{\textbf{Relation features}}\\
        \hline
        1 & Relation \\
        \hline
        2 & Has\_constant \\
        \hline
        3 & Constant value(s) \\
        \hline
        \multicolumn{2}{|c|}{\textbf{Partitioning features}}\\
        \hline
        4$_i$ & dim$_i$\_same \\
        \hline
        5$_{ij}$ & dim$_i$\_ldim$_j$\_same \\
        \hline
        \multicolumn{2}{|c|}{\textbf{Conditioning features}}\\
        \hline
        6$_i$ & dim$_i$\_avg\_dist \\
        \hline
        7$_{ij}$ & dim$_i$\_ldim$_j$\_avg\_dist \\
        \hline
    \end{tabular}
\end{table}

\textbf{Relation features}
The first feature represents the constraint's relation, such as equality, inequality, or a specific function name. The second feature indicates whether at least one constant value is present in the constraint. The remaining relation features represent numerical values of the constants. Since the list of features must be of fixed size, we include as many features for constants as the largest number of constants present in any constraint from the input set of constraints $C_A$. For constraints with fewer constants, the unnecessary features get ``NaN'' values.

\textbf{Partitioning features}
In many cases, the variables $V$ are given in the form of a matrix or tensor. The dimensions of such a tensor, and the index of each variable in them, often play a crucial role in the partitions within the problem. For example, the same constraint may be defined on all groups of variables that are part of the same row or column of tensor. Thus, we include information about the indices of the variables in these dimensions. We include as many index features as the largest number of dimensions over all variables, again using ``NaN'' values for constraints whose variables have fewer dimensions. 

Besides using only the dimensions present in the variables tensor, we also aim to identify ``latent dimensions'' that can be indirectly derived from the problem parameters $\mathcal{P}$. This is inspired by the \verb|scheme| sequence generator of ModelSeeker~\cite{beldiceanu2012model}.
For example, the blocks in a Sudoku grid form latent dimensions. A block index is not explicitly present in the variable matrix, but can easily be obtained by dividing the row/column index by the square root of the grid size.

To construct the latent dimensions, we examine each parameter value of the input problem instance(s) as a possible divisor of the variable tensor's dimension sizes. We add as many latent dimensions as there are valid divisors for the different dimensions in the input problem instances.

\textbf{Conditioning features}
There may be sequence conditions that restrict which combinations of variables the constraints should apply to within the variable partition. The classifier must be able to detect these conditions, so they must be detectable in the feature representation. Note that partitioning features can also be used to capture the sequence conditions, since a condition may state that, for example, the variables must not be part of the same row in the variable matrix. However, there may also be more complex sequence conditions. To capture these, we include features that express the average difference of the variable indices in each dimension and latent dimension, among all variables in the scope of the given constraint.

\subsection{Benchmark details}

We used 4 benchmarks commonly used in the CA literature. We focused on using benchmarks that have different constraint specifications, so that our method is evaluated in distinct cases. We used the following benchmarks, with 10 instances in each benchmark:

\textbf{Sudoku}
The Sudoku puzzle is a $n \times n$ grid, which must be completed in such a way that all the rows, columns, and $n$ non-overlapping $b\times b$ blocks contain distinct numbers. We used a variation where the blocks do not need to be square, i.e., $br \times bc$ blocks, with $br$ being the height of the block and $br$ its width. The variables are the grid cells, arranged in a 2D matrix with domains the different values they can take. We considered as parameter the grid size and the block height and width, i.e., $\mathcal{P} = \{n, br, bc\}$. We generated 10 random different instances, with:
\begin{itemize}
    \item Grid size $n$: $4 <= n <= 16$, ensuring that $n$ has at least 1 divisor other than itself and 1.
    \item Block height $br$: $2 <= br <= \lfloor n/2 \rfloor$, with $n \mod br = 0$.
    \item block width $bc$: $bc = \lfloor n/br \rfloor$.
\end{itemize}

\textbf{Golomb rulers.}
The problem is to find a ruler with $m$ marks, where the distance between any two marks is different from that between any other two marks. The variables are the marks, arranged in a vector.
We considered as parameter the amount of marks, i.e., $\mathcal{P} = \{m\}$. We generated 10 instances with $5 <= m < 15$.

\textbf{Exam Timetabling}
There are $s$ semesters, each containing $\mathit{n}$ courses, and we want to schedule the exams of the courses in a period of $d$ days, On each day we have $t$ available timeslots. The variables are the courses ($|V| = ns \cdot \mathit{n}$), arranged in a 2D matrix, having as domains the timeslots they can be assigned on, i.e., ($D_i = 1, ..., r \cdot t \cdot d$).
The constraints define that all courses need to be scheduled in a different timeslot, while exams of courses from the same semester must be scheduled
on different days. Thus, there are $\neq$ constraints between each pair of exams and \verb|different_day| constraints between courses of the same semester.
The latter are expressed by constraints of the type
$\lfloor v_1/spd \rfloor \neq \lfloor v_2/spd \rfloor$. 
The parameters of the problem are $\mathcal{P} = \{s, \mathit{n}, d, t\}$. We generated 10 random instances, with:
\begin{itemize}
    \item Semesters $s$: $4 <= s <= 10$.
    \item Courses per semester $n$: $3 <= n <= 7$.
    \item Timeslots $t$: $2 <= t <= 12$.
    \item Days $d$: $days_{min} <= d <= days_{min}+5$, with $days_{min} = \lceil n\times s / t\rceil$ to ensure it is satisfiable.
\end{itemize}

\textbf{Nurse rostering}
There are $n$ nurses, $s$ shifts per day, $ns$ nurses per shift, and $d$ days. The goal is to create a schedule, assigning a nurse to all existing shifts. The variables are the shifts, and there are a total of $d \cdot s \cdot ns$ shifts. The variables are modeled in a 3D matrix. The domains of the variables are the nurses. 
Each shift in a day must be assigned to a different nurse and the last shift of a day must be assigned to a different nurse than the first shift of the next day. The parameters of the problem are $\mathcal{P} = \{n, s, ns, d\}$. 
\begin{itemize}
    \item Nurses $n$: $12 <= n <= 20$.
    \item Shifts per day $s$: $3 <= s <= 4$.
    \item Nurses per shift $ns$: $3 <= ns <= \lfloor n / s \rfloor$.
    \item Days $d$: $7 <= d <= 30$.
\end{itemize}

\subsection{Metrics}

For our evaluation, we have defined the following:
\begin{itemize}
    \item True Positives (TP): the constraints that were correctly identified as being present in the target instance
    \item False Positives (FP): the constraints that were incorrectly identified as being present in the target instance when they are not.
    \item False Negatives (FN): the constraints that are part of the target model, but were not identified.
\end{itemize}

Note that, for a constraint to have a positive ground truth label, our evaluation did not only check if a (candidate) constraint is part of the ground truth set of constraints but also if it is logically implied by it.

Using the defined concepts, our evaluation is based on the following metrics that are common in ML:

\textbf{Precision (Pr)}: Precision is a measure of the accuracy of the identified constraints in the target instance. A high precision score signifies a low rate of false positive errors, indicating that when the method identifies a constraint, it is likely to be correct. When the precision score is 100\%, the predicted set of constraints is sound. It is calculated as: 
\begin{equation}
Pr = \frac{TP}{TP + FP}
\end{equation}

\textbf{Recall (Re)}: Recall is a measure of the method's ability to identify all relevant constraints. 
A high recall score indicates a low rate of false negative errors, indicating that the method is effective in identifying all relevant constraints and has a low rate of false negatives. When the recall score is 100\%, the predicted set of constraints is complete. It is calculated as: 
\begin{equation}
Re = \frac{TP}{TP + FN}
\end{equation}

\subsection{Tuning details}
We tuned the most important hyperparameters for MLP and KNN. A grid search was conducted using all benchmarks utilized in the experiments, with consecutive instances being used for training and testing. We used balanced accuracy as the tuning metric to address class imbalance.

Regarding KNN, we explored different distance metrics (euclidean, manhattan, and minkowski), weights (uniform and distance) and number of neighbours (1-21). The optimal configuration used euclidean distance metric, uniform weights and 2 neighbours.

For the MLPs, we focused on tuning the number of hidden layers (1-4), the number of neurons per layer ([8, 16, 32, 64]), the optimizer (sgd and adam) and the learning rate ([0.001, 0.01, 0.1, 1]). The optimal configuration was identified as an MLP with a single hidden layer having 64 neurons, using adam as the optimizer and with a learning rate of 0.01, using ReLu as the activation function.

\subsection{Impact of noise per benchmark} 

\begin{figure}[tb]
    \centering
    \begin{subfigure}[b]{0.22\textwidth}
        \centering
        \includegraphics[width=\textwidth]{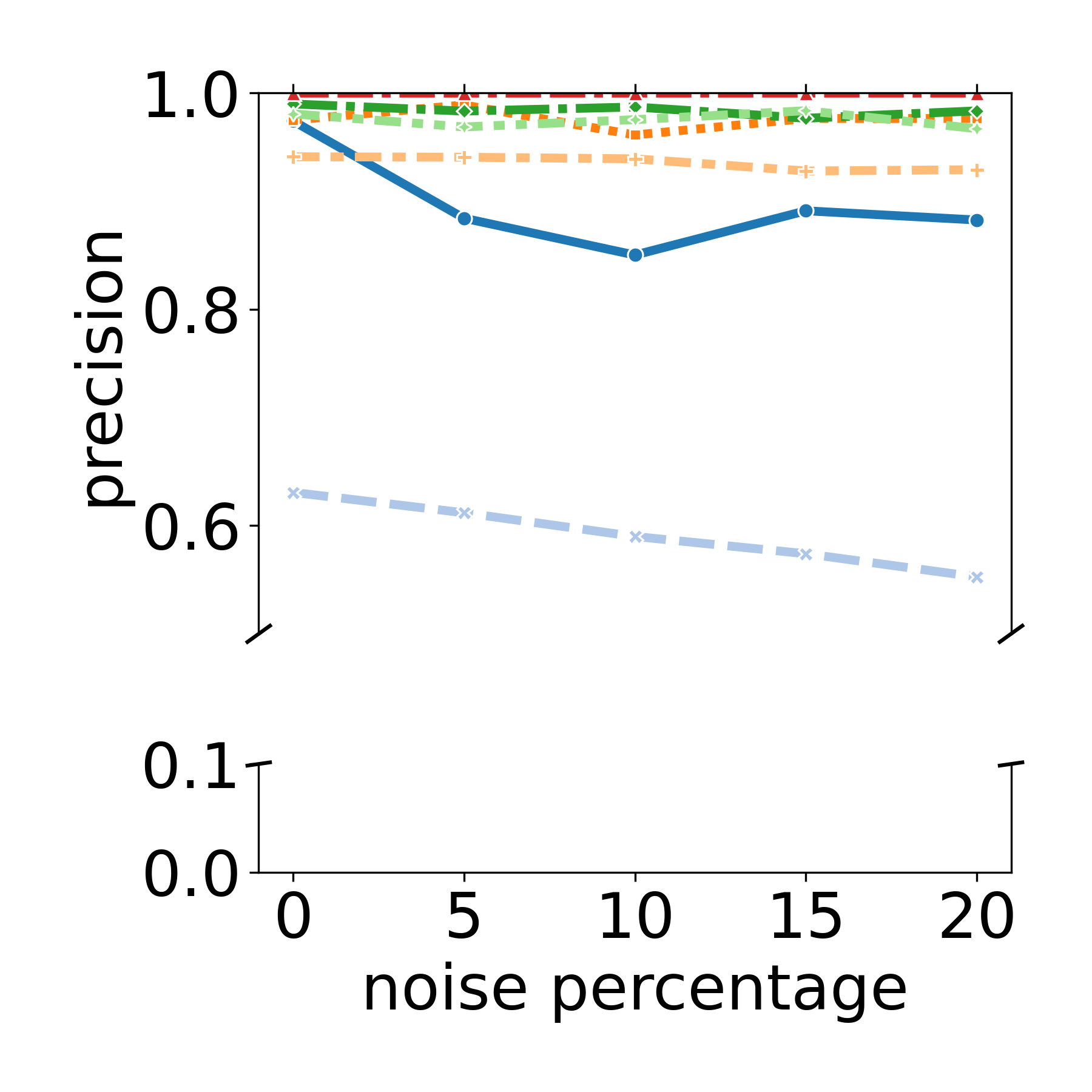}
        \caption{Precision with FP noise}
    \end{subfigure}
    \hfill
    \begin{subfigure}[b]{0.22\textwidth}
        \centering
        \includegraphics[width=\textwidth]{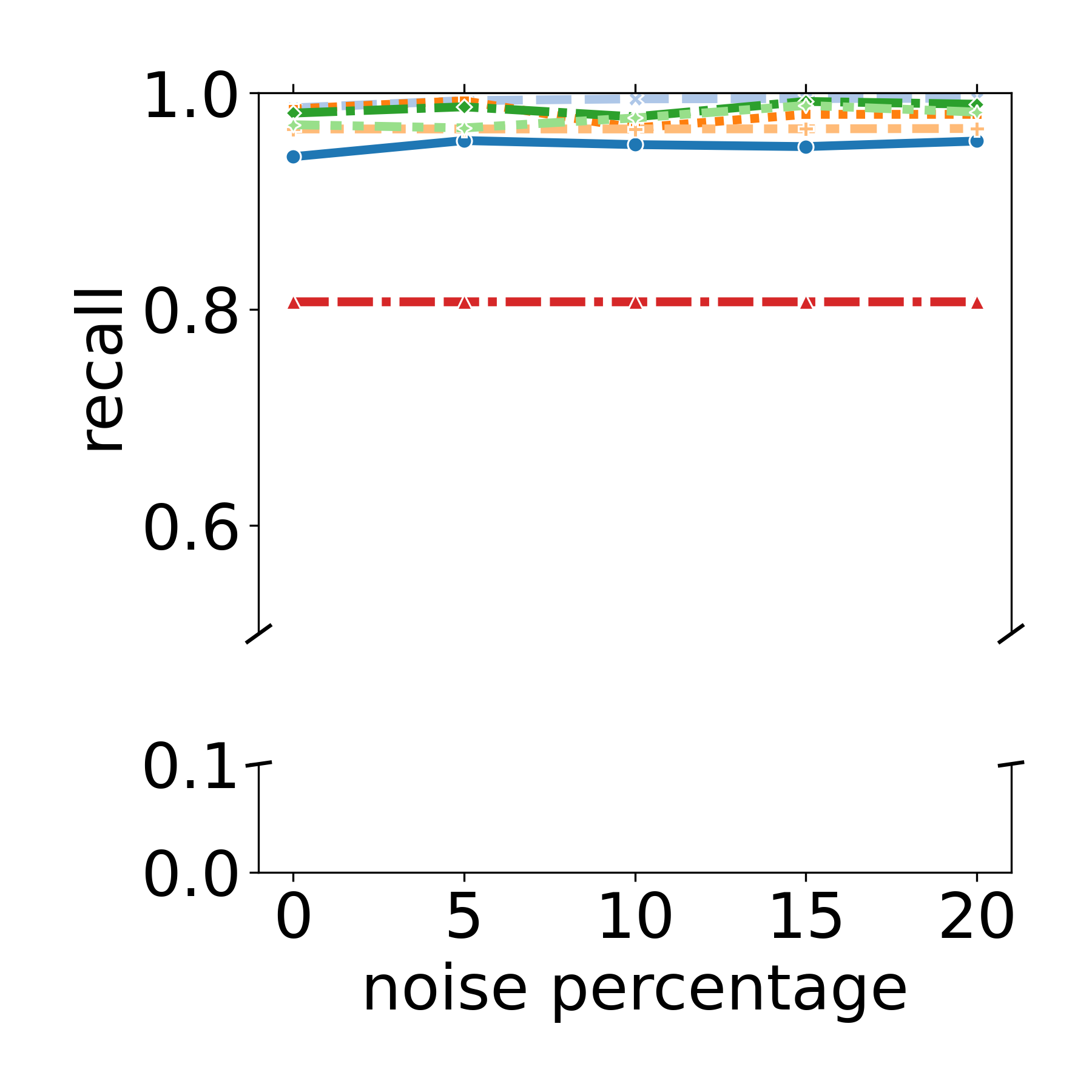}
        \caption{Recall with FP noise}
    \end{subfigure}
    \hfill
    \begin{subfigure}[b]{0.22\textwidth}
        \centering
        \includegraphics[width=\textwidth]{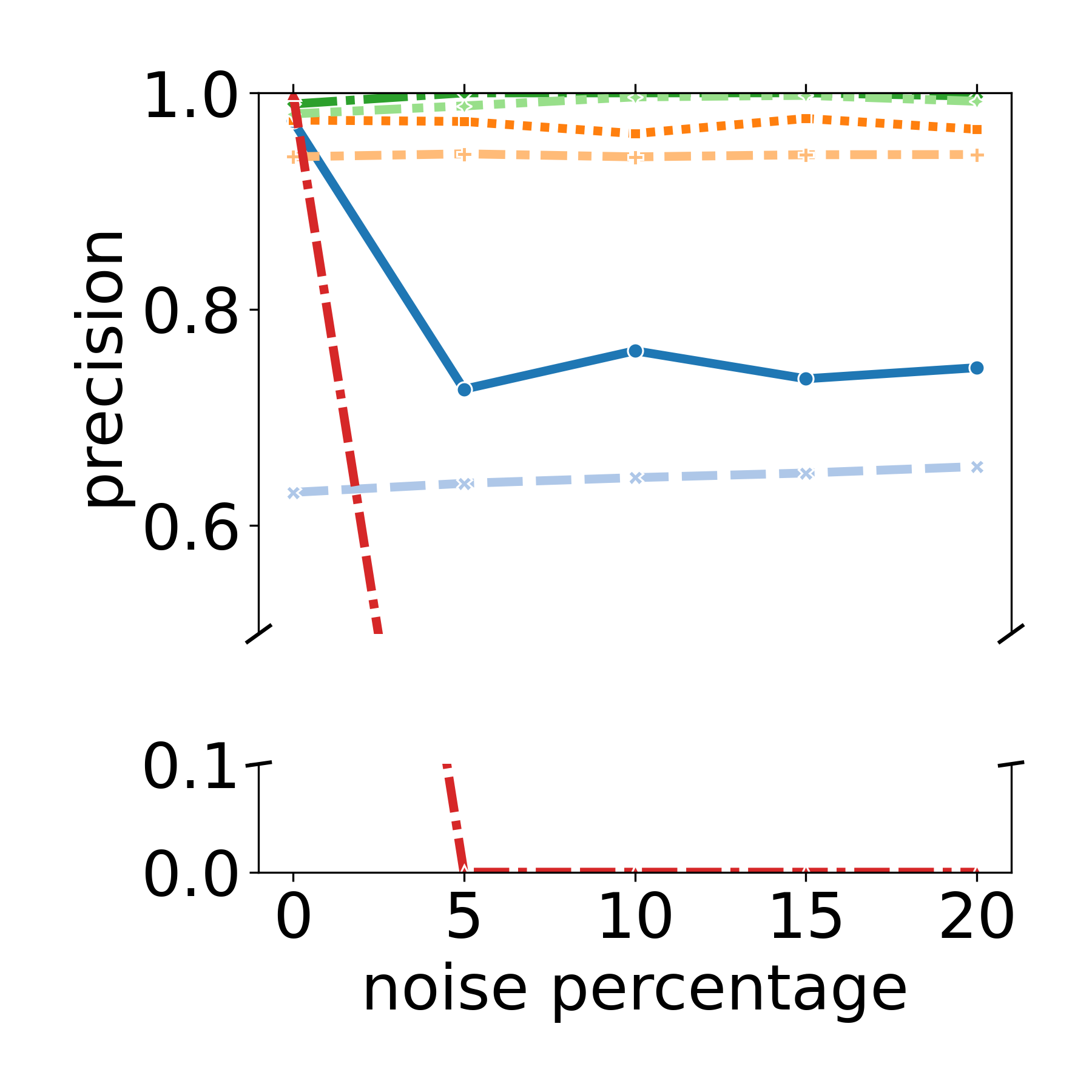}
        \caption{Precision with FN noise}
    \end{subfigure}
    \hfill
    \begin{subfigure}[b]{0.22\textwidth}
        \centering
        \includegraphics[width=\textwidth]{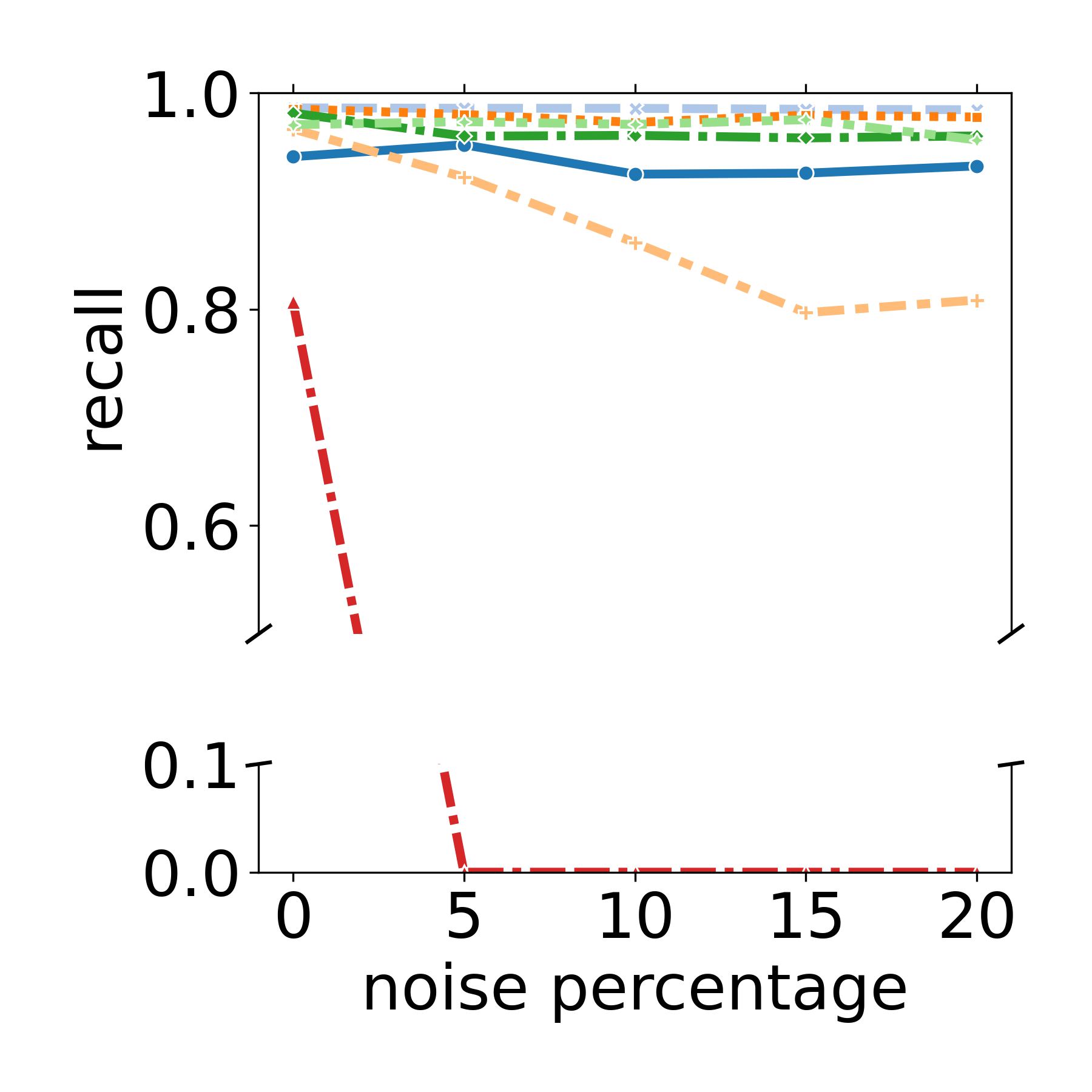}
        \caption{Recall with FN noise}
    \end{subfigure}
    \begin{subfigure}[b]{0.45\textwidth}
        \centering
        \includegraphics[width=\textwidth]{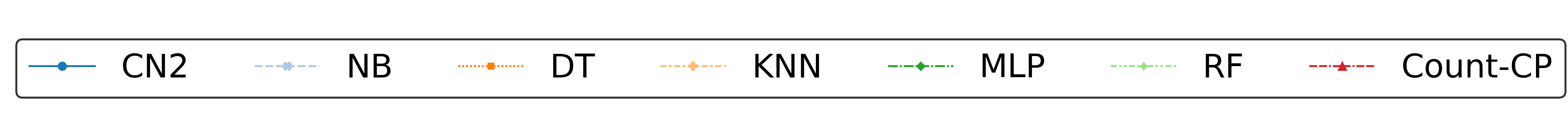}
    \end{subfigure}
    \caption{Results in Sudoku with the presence of noise}
    \label{fig:sudoku}
\end{figure}

\begin{figure}[tb]
    \centering
    \begin{subfigure}[b]{0.22\textwidth}
        \centering
        \includegraphics[width=\textwidth]{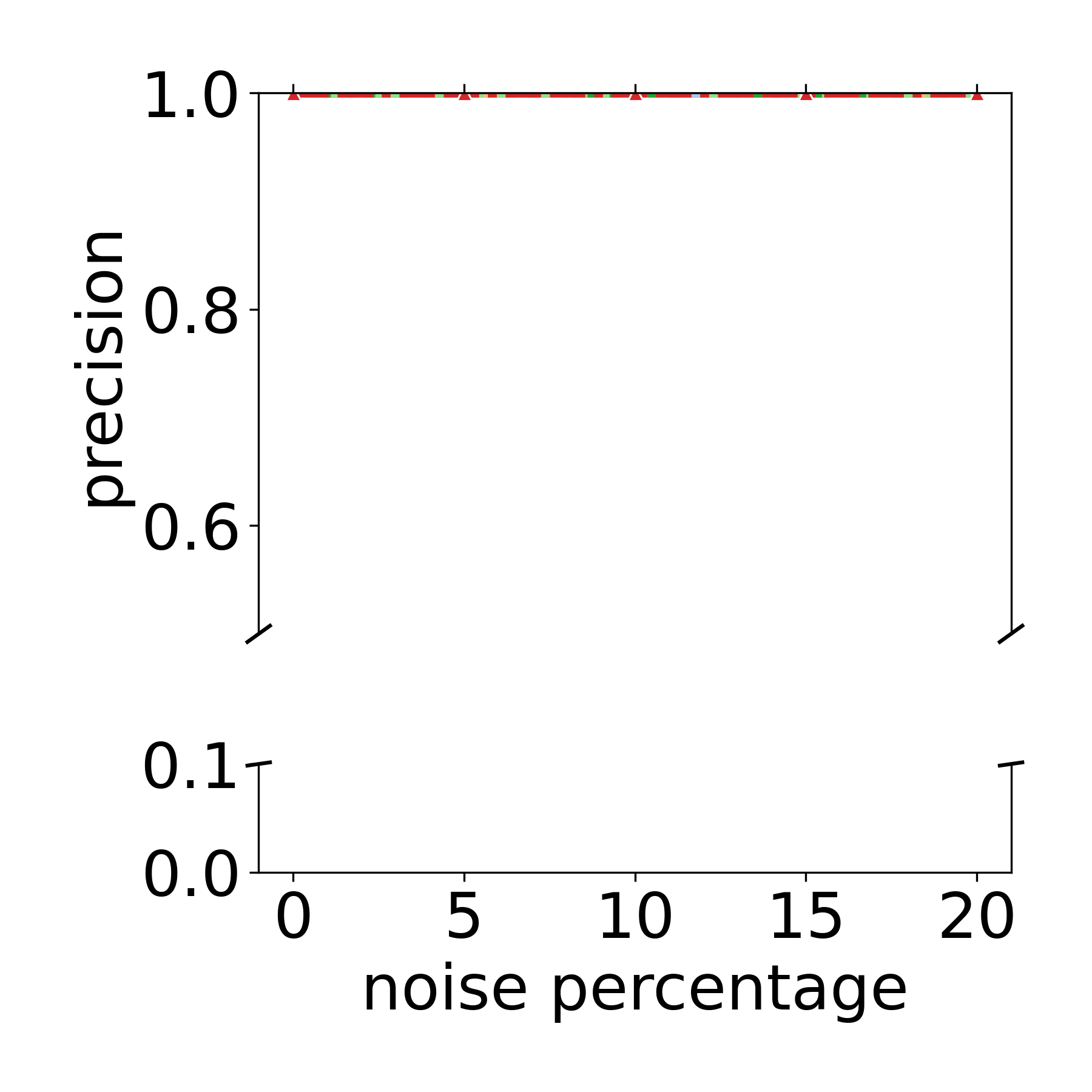}
        \caption{Precision with FP noise}
    \end{subfigure}
    \hfill
    \begin{subfigure}[b]{0.22\textwidth}
        \centering
        \includegraphics[width=\textwidth]{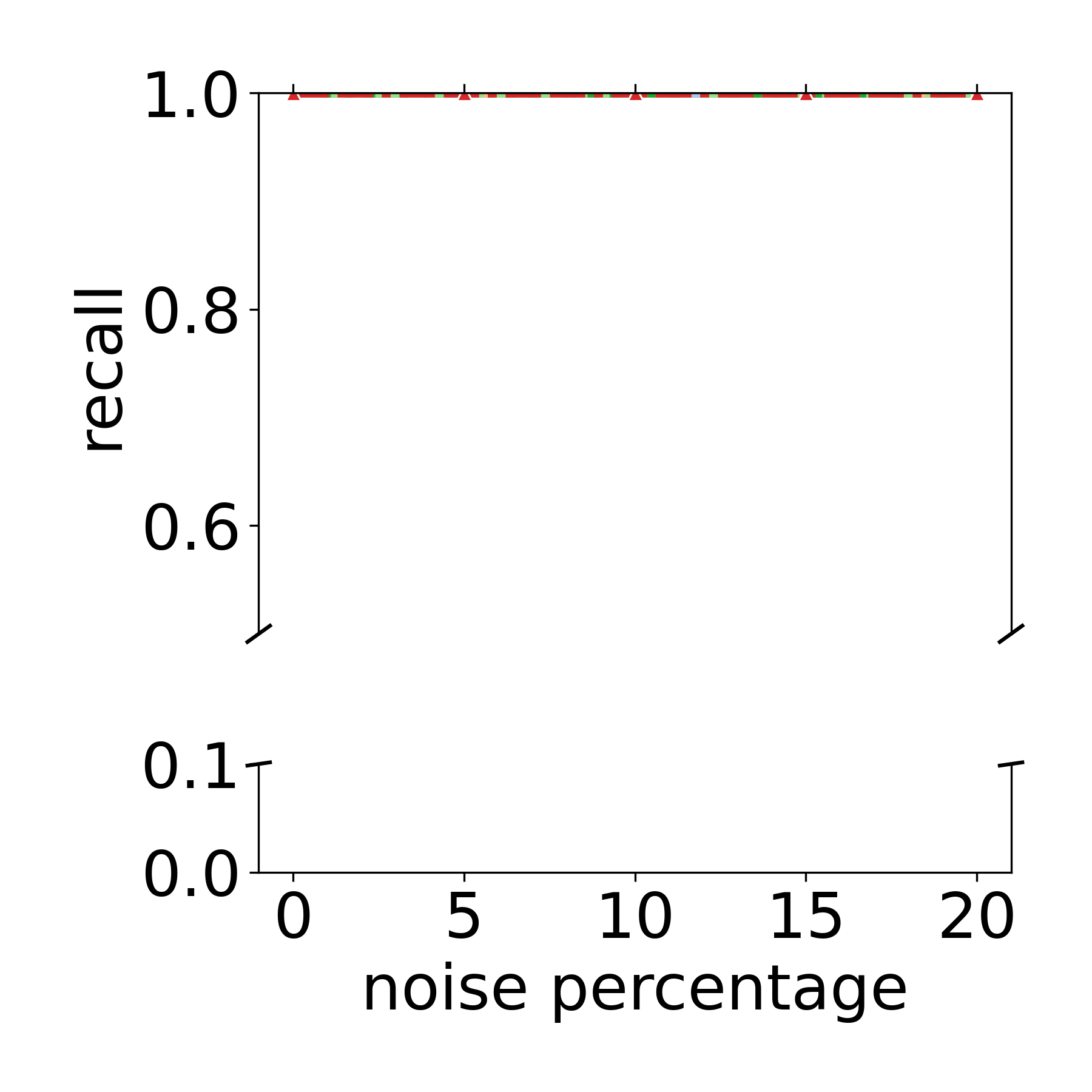}
        \caption{Recall with FP noise}
    \end{subfigure}
    \hfill
    \begin{subfigure}[b]{0.22\textwidth}
        \centering
        \includegraphics[width=\textwidth]{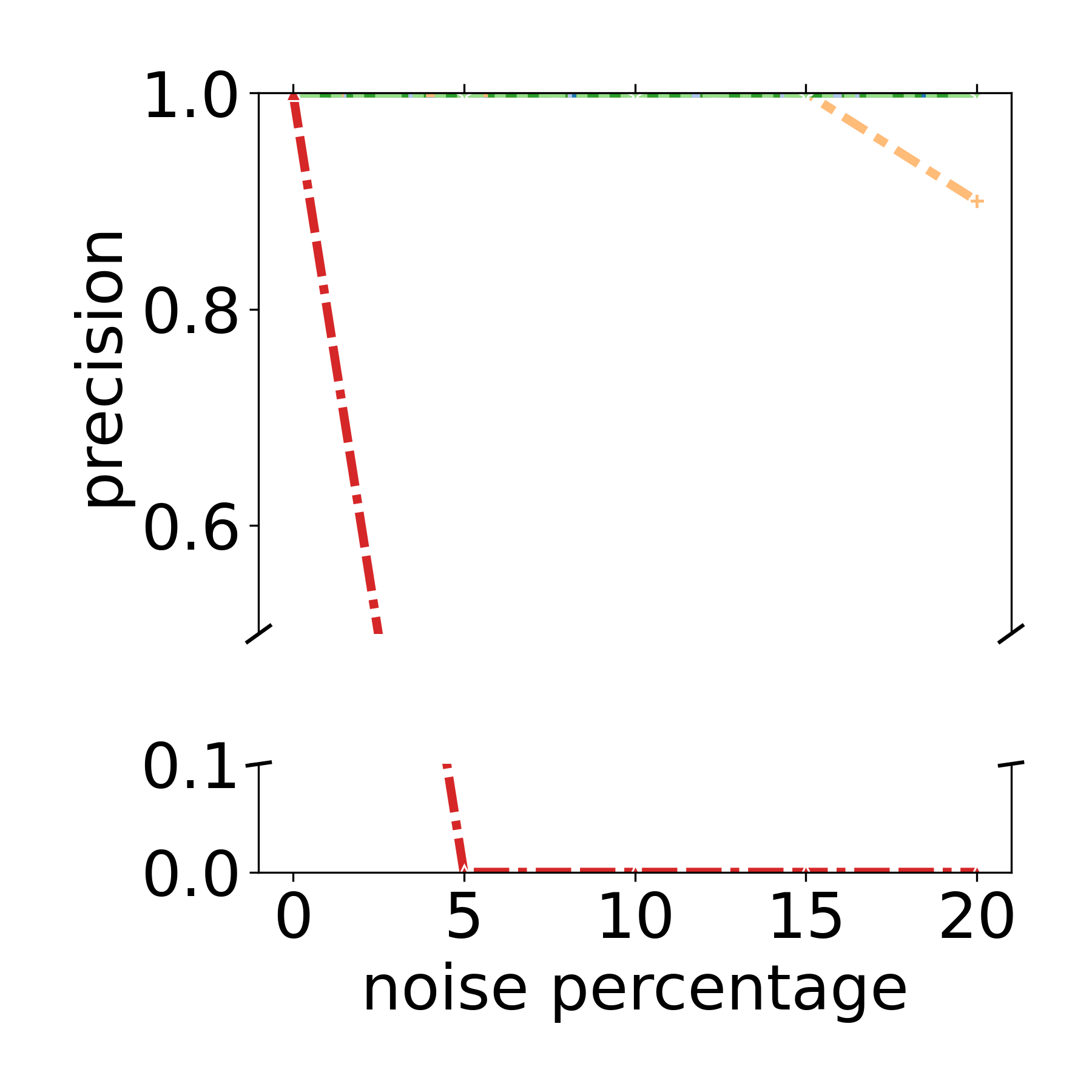}
        \caption{Precision with FN noise}
    \end{subfigure}
    \hfill
    \begin{subfigure}[b]{0.22\textwidth}
        \centering
        \includegraphics[width=\textwidth]{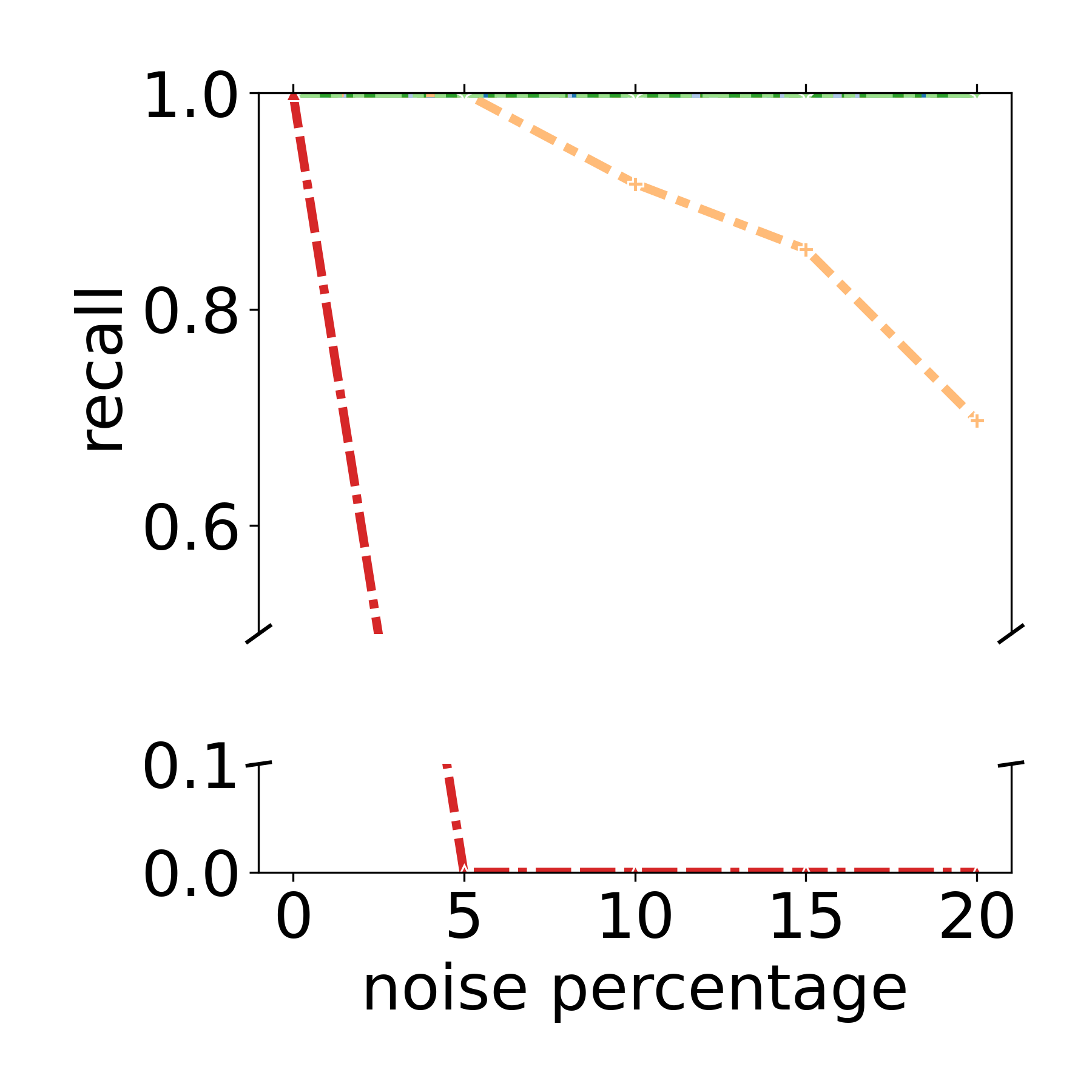}
        \caption{Recall with FN noise}
    \end{subfigure}
    \begin{subfigure}[b]{0.45\textwidth}
        \centering
        \includegraphics[width=\textwidth]{results/legend.png}
    \end{subfigure}
    \caption{Results in Golomb with the presence of noise}
    \label{fig:golomb}
\end{figure}

\begin{figure}[tb]
    \centering
    \begin{subfigure}[b]{0.22\textwidth}
        \centering
        \includegraphics[width=\textwidth]{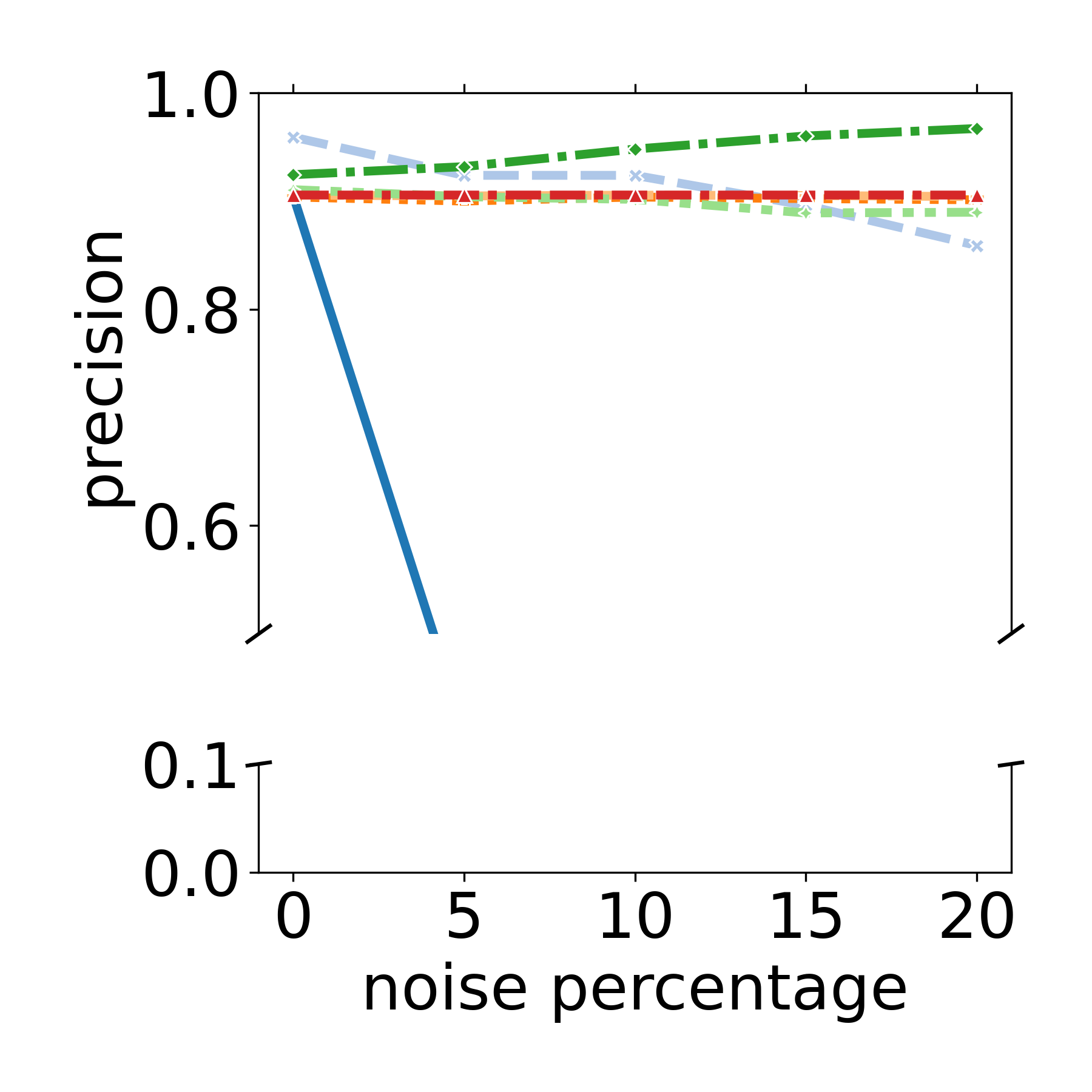}
        \caption{Precision with FP noise}
    \end{subfigure}
    \hfill
    \begin{subfigure}[b]{0.22\textwidth}
        \centering
        \includegraphics[width=\textwidth]{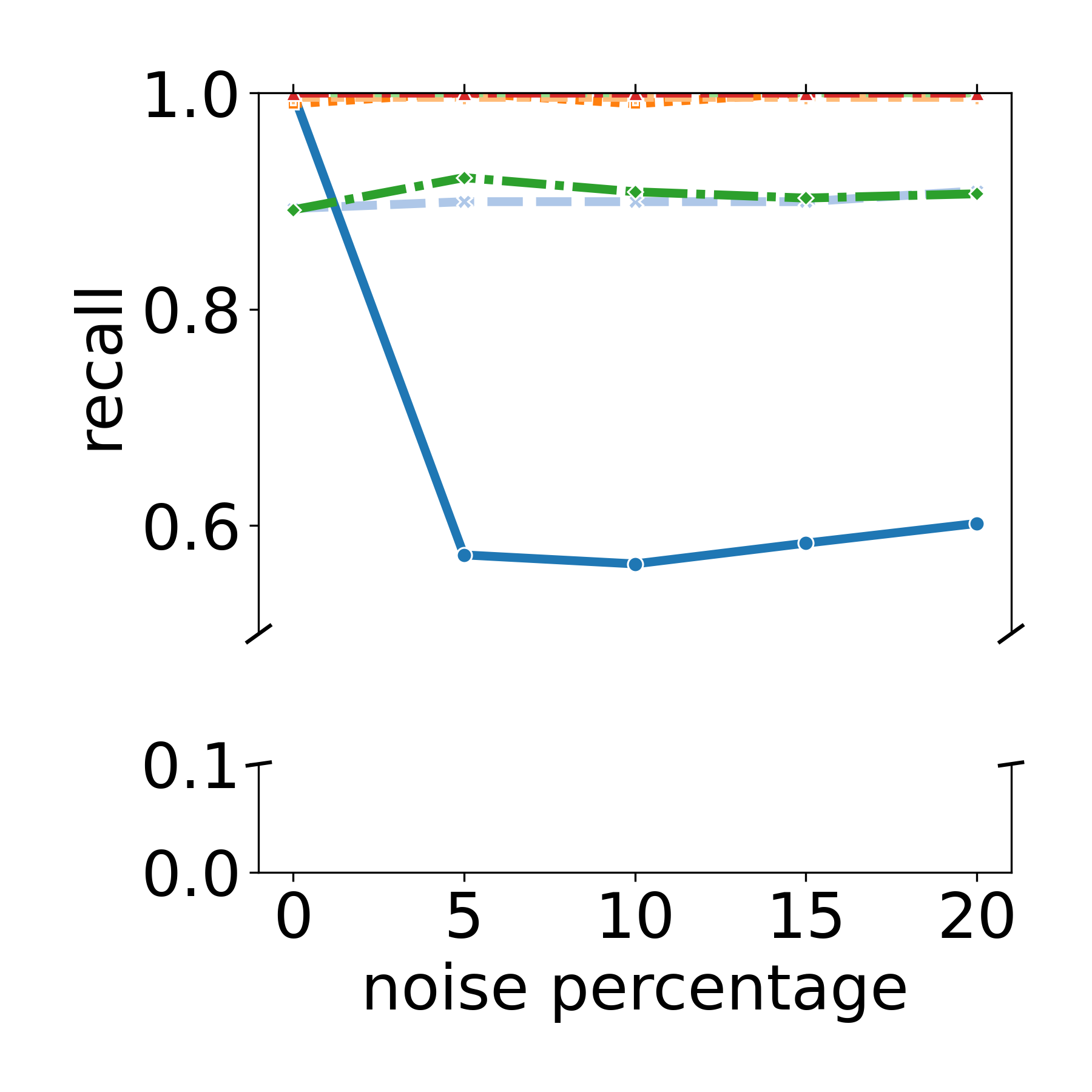}
        \caption{Recall with FP noise}
    \end{subfigure}
    \hfill
    \begin{subfigure}[b]{0.22\textwidth}
        \centering
        \includegraphics[width=\textwidth]{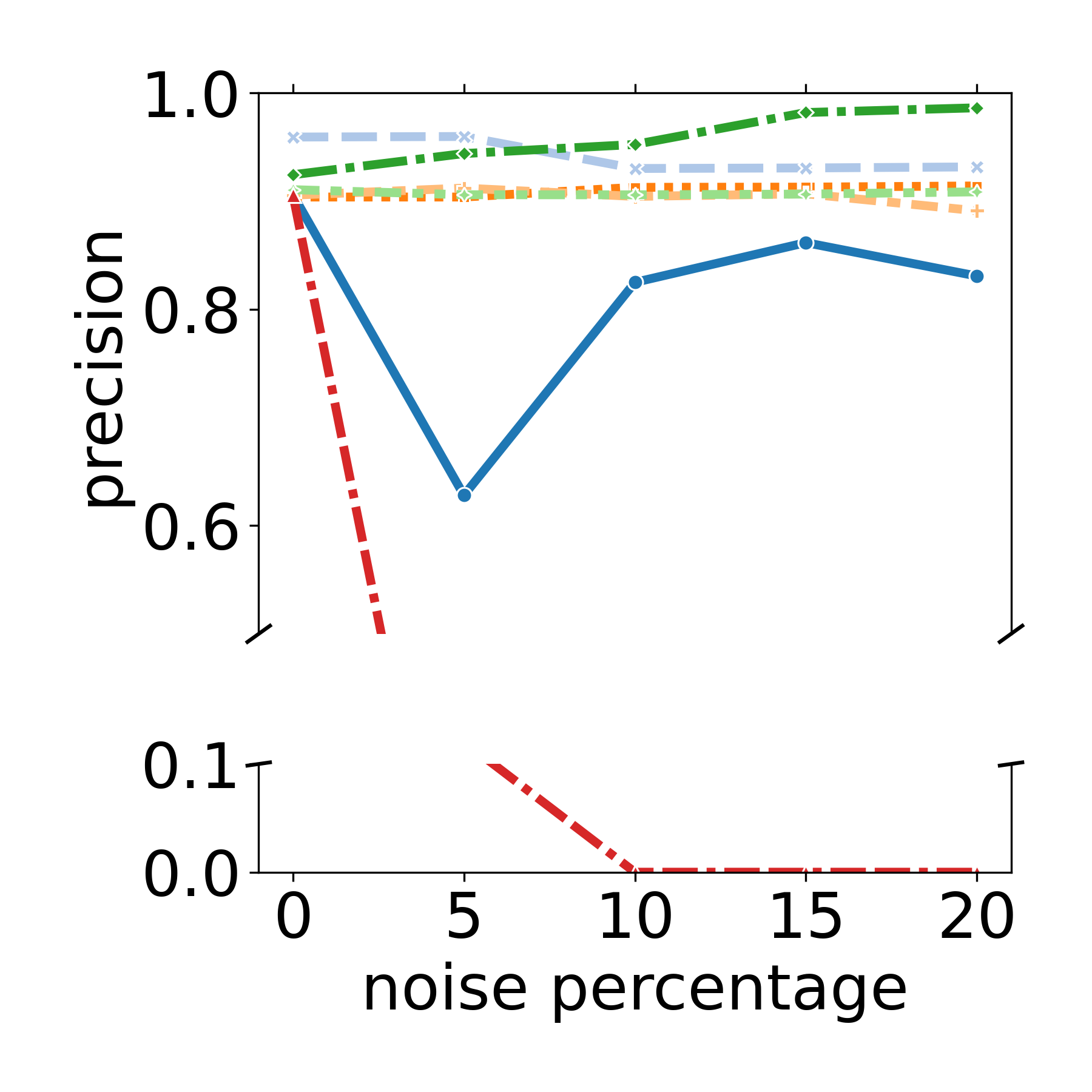}
        \caption{Precision with FN noise}
    \end{subfigure}
    \hfill
    \begin{subfigure}[b]{0.22\textwidth}
        \centering
        \includegraphics[width=\textwidth]{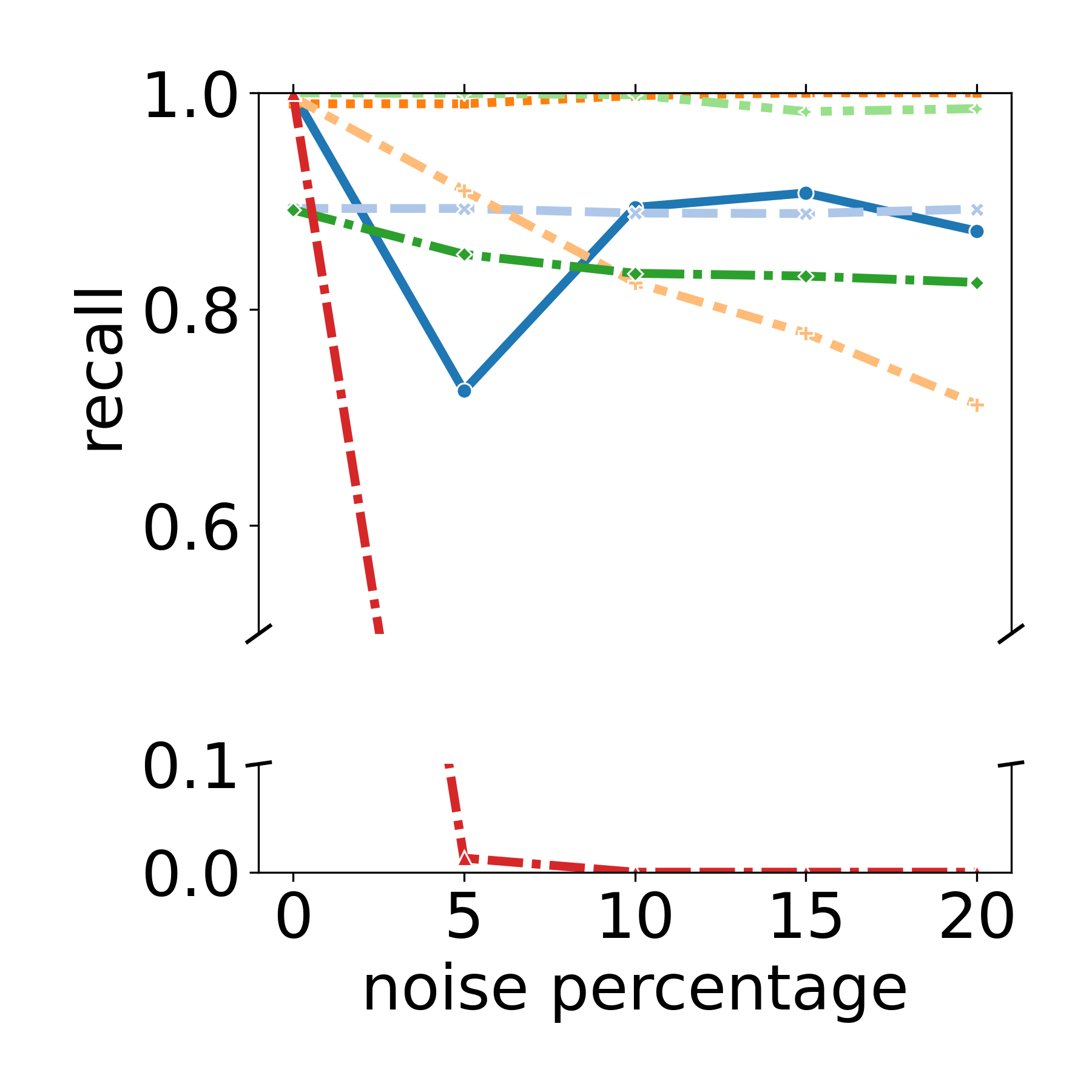}
        \caption{Recall with FN noise}
    \end{subfigure}
    \begin{subfigure}[b]{0.45\textwidth}
        \centering
        \includegraphics[width=\textwidth]{results/legend.png}
    \end{subfigure}
    \caption{Results in ET with the presence of noise}
    \label{fig:et}
\end{figure}

\begin{figure}[tb]
    \centering
    \begin{subfigure}[b]{0.22\textwidth}
        \centering
        \includegraphics[width=\textwidth]{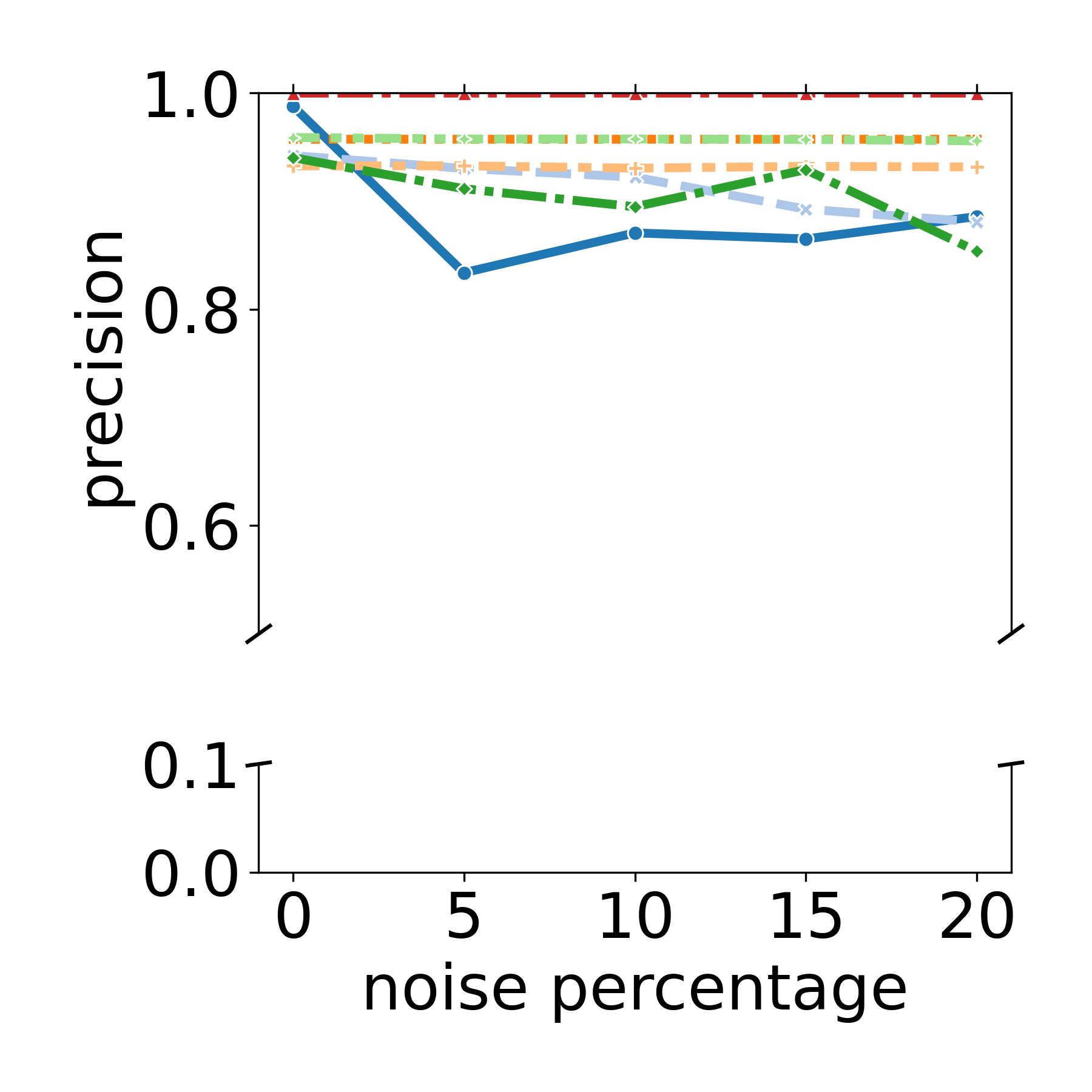}
        \caption{Precision with FP noise}
    \end{subfigure}
    \hfill
    \begin{subfigure}[b]{0.22\textwidth}
        \centering
        \includegraphics[width=\textwidth]{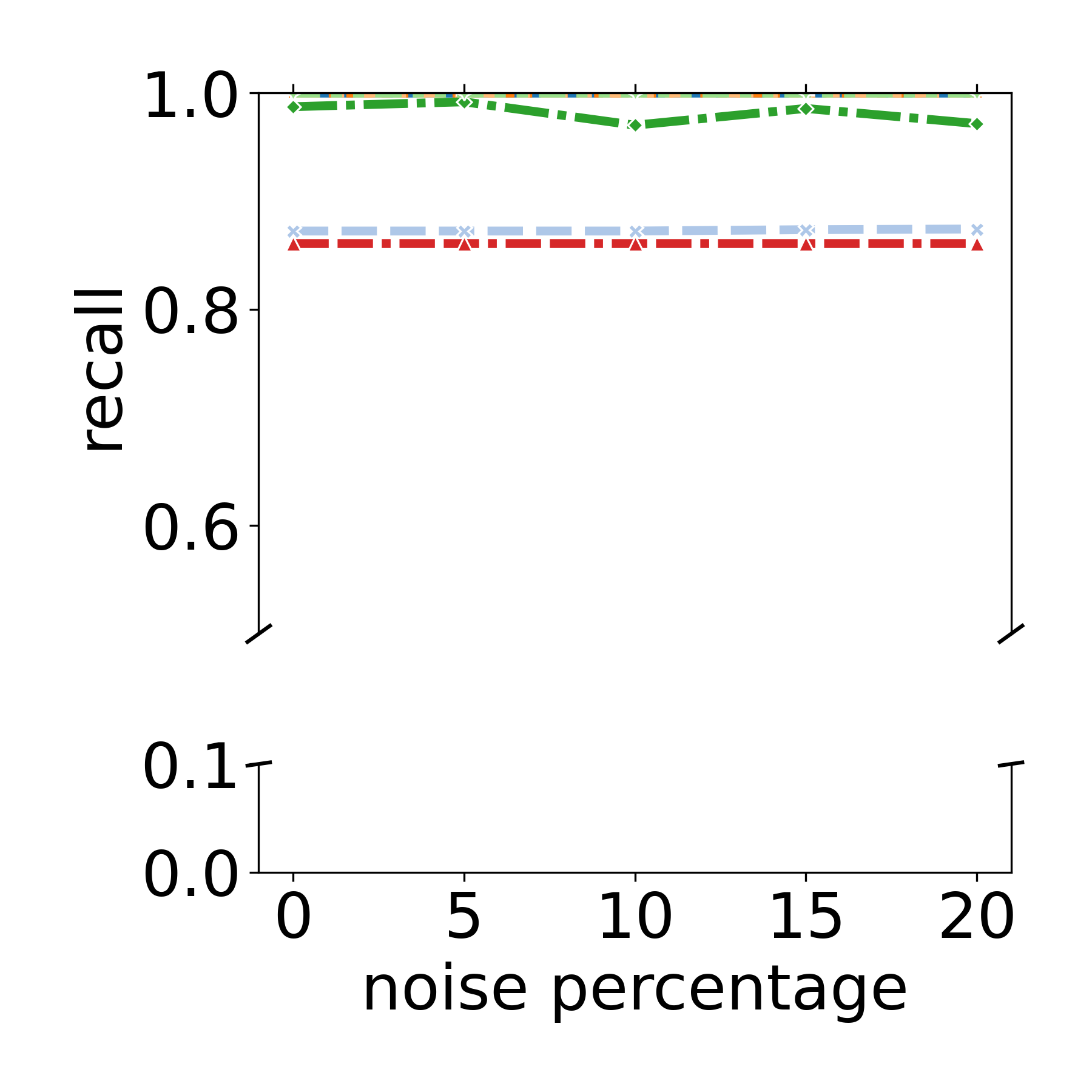}
        \caption{Recall with FP noise}
    \end{subfigure}
    \hfill
    \begin{subfigure}[b]{0.22\textwidth}
        \centering
        \includegraphics[width=\textwidth]{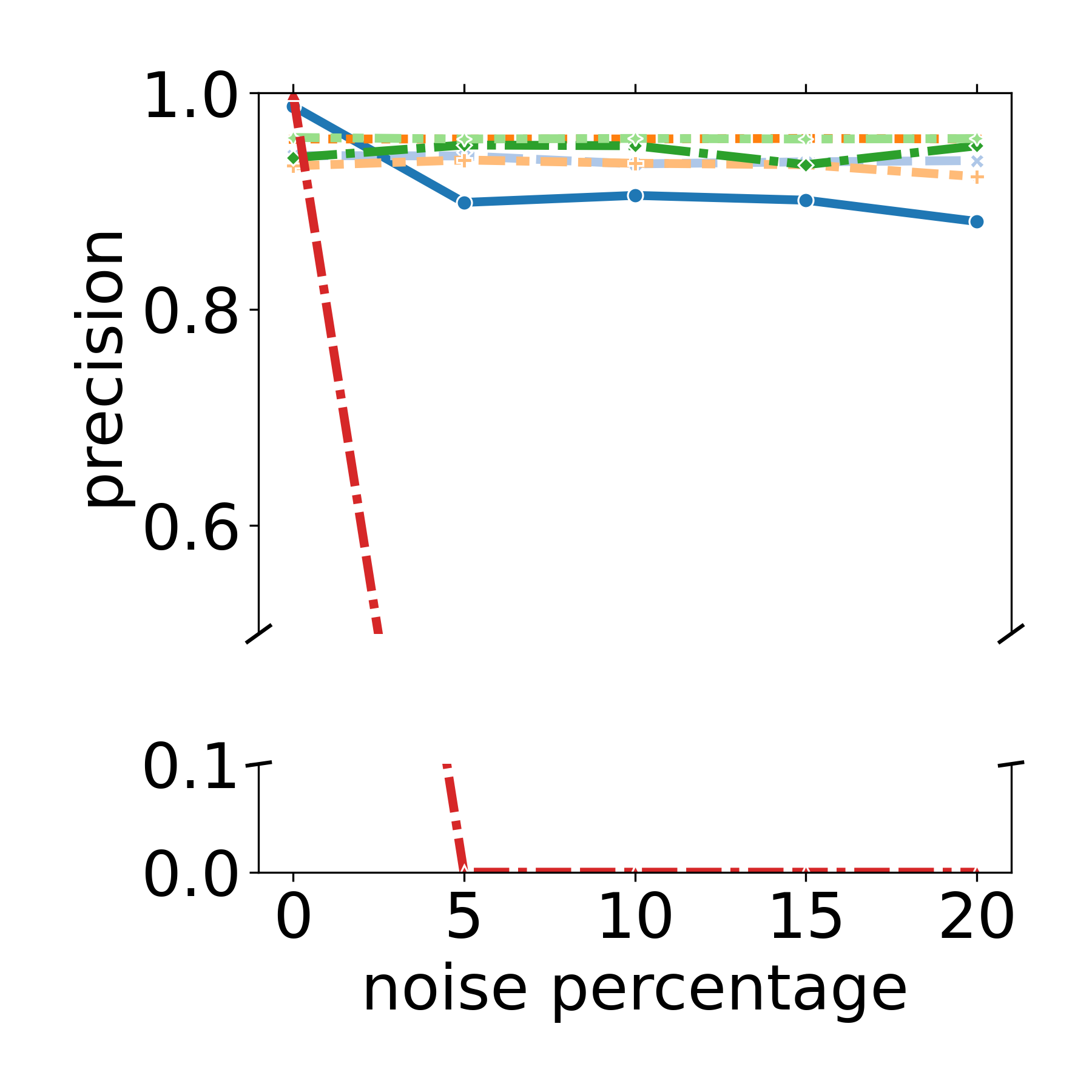}
        \caption{Precision with FN noise}
    \end{subfigure}
    \hfill
    \begin{subfigure}[b]{0.22\textwidth}
        \centering
        \includegraphics[width=\textwidth]{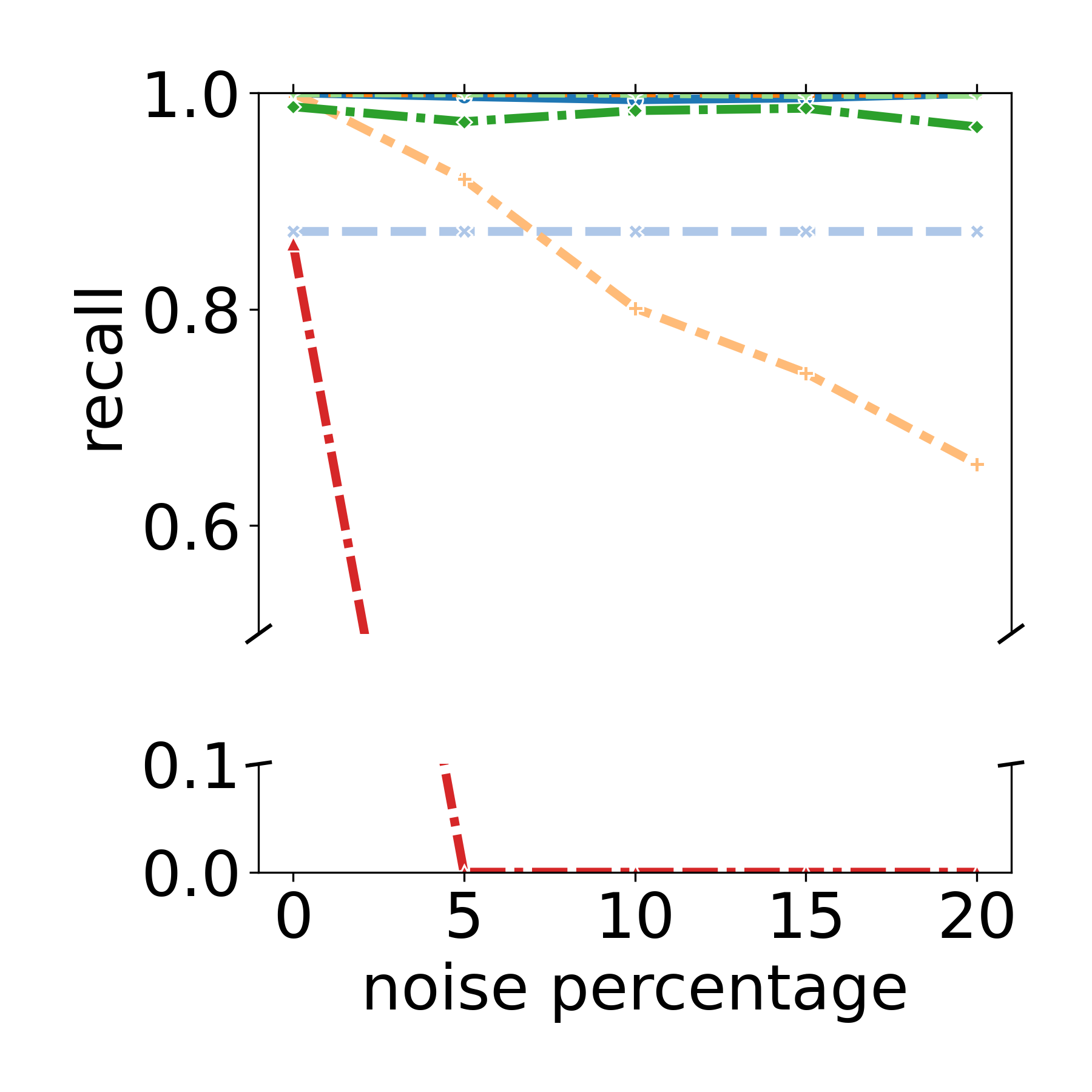}
        \caption{Recall with FN noise}
    \end{subfigure}
    \begin{subfigure}[b]{0.45\textwidth}
        \centering
        \includegraphics[width=\textwidth]{results/legend.png}
    \end{subfigure}
    \caption{Results in NR with the presence of noise}
    \label{fig:nr}
\end{figure}

Figures~\ref{fig:sudoku}-\ref{fig:nr} present the detailed results of our method (using various classifiers) compared to \textsc{Count-CP} in the presence of noise.

\newpage

\end{document}